\documentclass[3p,12pt,english]{elsarticle}

\usepackage{amsmath,amssymb,amsfonts}
\usepackage{booktabs}
\usepackage{algorithmic}
\usepackage{graphicx}
\usepackage{textcomp}
\usepackage{xcolor}
\usepackage{comment}
\usepackage{url}
\usepackage[linesnumbered,ruled,vlined]{algorithm2e}
\usepackage{csvsimple}
\usepackage{varwidth}
\usepackage{graphicx,wrapfig,lipsum}
\usepackage{tabularx}
\usepackage{csquotes}
\usepackage{array}
\usepackage{amsthm,epstopdf}
\usepackage{pgf}
\usepackage{natbib}
\newtheorem{defn}{Definition}

\def\BibTeX{{\rm B\kern-.05em{\sc i\kern-.025em b}\kern-.08em
    T\kern-.1667em\lower.7ex\hbox{E}\kern-.125emX}}

\makeatletter
\newcommand{\mypm}{\mathbin{\mathpalette\@mypm\relax}}
\newcommand{\@mypm}[2]{\ooalign{%
		\raisebox{.1\height}{$#1+$}\cr
		\smash{\raisebox{-.6\height}{$#1-$}}\cr}}
\makeatother

\newlength\mylen
\newcommand\myinput[1]{%
	\settowidth\mylen{\KwIn{}}%
	\setlength\hangindent{\mylen}%
	\hspace*{\mylen}#1\\}

\newcommand{\LineNumbered}{%
	\setboolean{algocf@linesnumbered}{true}%
	\renewcommand{\algocf@linesnumbered}{\everypar={\nl}}}%

\let\oldnl\nl% Store \nl in \oldnl
\newcommand{\nonl}{\renewcommand{\nl}{\let\nl\oldnl}}% Remove line number for one line

\graphicspath{{figs/}{figs/Paper3figs/}{figs/Paper1figs/}{figs/Paper2figs/}{figs/Paper3figs/}{figs/other/}{/other/}}

\begin{document}

	\begin{frontmatter}
		
		\title{LoRMIkA: Local Rule-based Model Interpretability with k-optimal Associations}

		%%%%%New

		%\author{}
		\author[monash1]{Dilini~Rajapaksha \corref{cor1}}%\corref{cor1}
		\author[monash1]{Christoph~Bergmeir}
		\author[monash1]{Wray~Buntine}

		\address{Dilini.Rajapakshahewaranasinghage@monash.edu, Christoph.Bergmeir@monash.edu, Wray.Buntine@monash.edu}
		\address[monash1]{Faculty of Information Technology, Monash University, Melbourne, Australia.}
		%	\address[uber]{Uber Technologies Inc, San Francisco, California, United States.}

		\cortext[cor1]{Postal Address: Faculty of Information Technology, P.O. Box 63 Monash University, Victoria 3800, Australia. E-mail address: Dilini.Rajapakshahewaranasinghage@monash.edu}
		\cortext[cor2]{Postal Address: Faculty of Information Technology, P.O. Box 63 Monash University, Victoria 3800, Australia. E-mail address: Christoph.Bergmeir@monash.edu}
		\cortext[cor2]{Postal Address: Faculty of Information Technology, P.O. Box 63 Monash University, Victoria 3800, Australia. E-mail address: Wray.Buntine@monash.edu }
		%	\cortext[cor3]{Postal Address: Uber Technologies, 1455 Market St 400, San Francisco, CA 94103, USA. E-mail address: Slawek@uber.com}

		\begin{abstract}
		As we rely more and more on machine learning models for real-life
		decision-making, being able to understand and trust the predictions
		becomes ever more important. Local explainer models have recently been
		introduced to explain the predictions of complex machine learning
		models at the instance level. In this paper, we propose Local
		Rule-based Model Interpretability with k-optimal Associations
		(LoRMIkA), a novel model-agnostic approach that obtains k-optimal
		association rules from a neighbourhood of the instance to be
		explained. Compared with other rule-based approaches in the literature,
		we argue that the most predictive rules are not necessarily the rules
		that provide the best explanations. Consequently, the LoRMIkA
		framework provides a flexible way to obtain predictive and interesting
		rules. It uses an efficient search algorithm guaranteed to find the
		k-optimal rules with respect to objectives such as confidence, lift, leverage, coverage, and support. It also provides multiple rules which
		explain the decision and counterfactual rules, which give
		indications for potential changes to obtain different outputs for
		given instances. We compare our approach to other
		state-of-the-art approaches in local model interpretability on three
		different datasets and achieve competitive results in terms
		of local accuracy and interpretability.
		\end{abstract}
		
		
	\end{frontmatter}

\section{Introduction}

%%\textbf{TODO: Revise 4.2 to end}
%
%\textbf{TODO: Connect the notation section better to the rest}
%
%\textbf{TODO: Give a better overview of the algorithm at the beginning of section 3.}
%
%\textbf{TODO (maybe): Explain better our error measures, what they are measuring and why this is important.}
%
%\textbf{TODO (maybe): 3.4: What is this good for? What can we do with these 4 different types of rules?}

Explainability of machine learning models is becoming ever more important~\citep{LiptonZachary2018-th}. For example, the European General Data Protection
Regulation from 2018 contains a {\em right to explanation} concept for any decision provided by predictive models. 
Traditionally, explainability can be achieved by machine learning models that are considered interpretable, such as classic linear regression, logistic regression, or decision trees~\citep{quinlan1986induction}. Also newer developments in this area of globally interpretable models are present, e.g. \citet{Proenca2019-nm} introduced an approach to formalize multi-class classification problems based on a probabilistic rule list and the minimum description length principle.  
However, recently local model-agnostic interpretability methods have been introduced that offer explanations for predictions of any (black-box) model. 
These methods typically use simpler models fitted locally in the neighbourhood of an instance to the output of a more complex black-box model, assuming that the behaviour of the instance to be explained is similar to the behaviour of its neighbourhood. In this way, they are able to provide \emph{local} explanations for a particular instance that can be more accurate and relevant for the instance under consideration as opposed to global explanations~\citep{Cano2013-gh,Azmi2019-rg} that do not change across the dataset.

The first work in this line that paved the way for others was the work of~\citet{ribeiro2016should} that introduced Local Interpretable Model-agnostic Explanations (LIME).
Here, instances are randomly perturbed around the instance to be explained and a linear model is fitted to the data in the selected neighbourhood of the considered instance. 
However, LIME has various limitations.  For instance, it does not perform well when the features have a higher degree of interaction and non-linear relationships with the target variable. Consequently,~\citet{Lundberg2017-hc} introduced SHapley Additive exPlanation (SHAP) that overcomes many of the shortcomings of LIME by providing consistent additive explanations based on game theory.
Both LIME and SHAP are feature-importance based explanations that show the features which have the most positive or negative impact on the prediction of the global model for a particular instance. %The main disadvantage of feature-importance based explainers is the inability to explain the prediction with logical conditions, which help to make decisions more easily.    

In recent years, deep learning methods have proven extremely successful, in particular in application areas where there are many highly dependent features like images, text, or speech. Here, considering the importance of features, i.e., pixels of an image or the words in a sentence, can often not be considered interpretable.
Researchers in these areas are addressing this issue mainly in two different ways. 
The first way is to use sample-based explanation methods that explain how the model parameters are actually derived based on the training data. For example, \citet{Koh2017-ln} considered influence functions to capture the core idea of the model in consideration. 
\citet{Yeh2018-vm} proposed a method to provide explanations for the prediction of a deep neural network for a given test instance, based on the representer theorem, which uses so-called representer values that measure the weighted importance of each training instance towards the learned parameters of the model. 
In their work, those authors discussed that the pre-activation prediction values can be decomposed into a linear combination of training point activations, with the weights corresponding to the representer values. 
The representer instances can be selected based on the magnitude of the representer values, which enables the understanding of the model prediction of the decomposition by identifying its representer instances in the training set. 
\citet{Khanna2018-ky} use Sequential Bayesian
Quadrature (SBQ) for efficient selection of instances and for feature embedding for each data point using Fisher kernels. \citet{DBLP:journals/corr/abs-1902-06292} apply an attention-based mechanism to identify the training set instances that relate to the prediction of the given test instance. 
The second main avenue for explainers in deep learning are concept-based explanations. For example, \citet{Been2018-vm} proposed Concept Activation Vectors (CAV) to interpret the internal state of the neural networks. The CAVs in this approach are considered to be human-friendly explanatory concepts. Further, \citet{Zhang2018-yw} considered multiple Gaussian models to represent the distribution of data where each Gaussian model reflects some local characteristics related to the dataset.
 
However, especially outside of the field of deep learning, recently many researchers have shifted from linear models as explainers to rule-based explanations~\citep{Proenca2019-nm,NOWAKBRZEZINSKA2019301}, as they arguably provide more precise explanations to the end users~\citep{Ribeiro2018-mc} and are more interpretable~\citep{Lakkaraju2016-rf} compared with others. \citet{DBLP:journals/corr/PuriGAVK17} introduced a global rule-based explainer.
\citet{Lakkaraju2019-kh} proposed a model-agnostic framework called Model Understanding through Subspace Explanations (MUSE), which explains the global model predictions by considering the different subgroups of the instances which are characterized by features of user interest.
\citet{Ribeiro2018-mc} proposed Anchor-LIME, which is a local-explainable model. It selects a neighbourhood with the help of a bandit algorithm~\citep{Kaufmann2013-bn} and extracts rules afterwards from the neighbourhood while finding subsets of features that retain the same prediction when held constant, even though all other features are changed. The rules obtained from that algorithm are called anchors by those authors.
A limitation of Anchor-LIME is that it does not provide counterfactual rules which are often important to invert a decision.
LOcal Rule-based Explanations (LORE) was introduced by \citet{DBLP:journals/corr/abs-1805-10820}. In this algorithm a decision tree is built in the neighbourhood of an instance to be explained, to acquire a single rule for the decision and a set of counterfactual rules for the inverse decision. 

In this paper, we propose a flexible framework for Local Rule-based Model Interpretability
with k-optimal Associations (LoRMIkA) to explain the predictions produced for tabular classification datasets.
Here, k is a user-defined parameter of the maximum number of (non-redundant) class association rules to be extracted. 

In accordance with other local-interpretability frameworks~\citep{ribeiro2016should,Ribeiro2018-mc,DBLP:journals/corr/abs-1805-10820}, the key assumption of LoRMIkA is that the behaviour of the instance to be explained is similar to the behaviour of the instances in its neighbourhood. 
To explain the instances in the neighbourhood, we first select the most similar instances to the instance to be explained from the training data set, and we generate synthetic instances within the neighbourhood. 
After, we generate the predictions for all the instances (i.e., both selected
instances and generated instances) in the neighbourhood using the global machine learning model to identify the logics and the behaviour of the global model when providing predictions. 
Here, we assume that the logic used to generate these predictions is similar to the prediction of the instance to be explained. 
Then we mine a total number of ``k'' optimal class association rules~\citep{webb2011filtered} between all the instances in the neighbourhood and their predictions using the OPUS search algorithm~\citep{Webb1995-cu} with respect to measures popular in the association rule mining community as described in Section~\ref{sec:assoc}.  

LoRMIkA leverages decades of research in association rule mining, and is therewith able to overcome a number of drawbacks of the existing algorithms. We argue that class association rules are better suited to provide explanations than, for instance, linear models or decision trees, as they are local models both with respect to features and instances. Furthermore, we consider one of the main limitations of current state-of-the-art rule-based local model-agnostic explainers to be that these algorithms consider only the most predictive rules. 
We argue that the most predictive rules are not necessarily the rules that explain the best. There can be \emph{interesting} rules which are very useful explainers though they are not highly predictive. The state-of-the-art algorithms do not consider such rules as the explanations of a prediction. We provide a detailed explanation of rules that predict vs rules that explain in Section~\ref{SubSec:Mining intersting rules}.
LoRMIkA uses an efficient search algorithm to find the k-optimal rules from a neighbourhood and is therewith able to produce model-agnostic explanations for a given black-box algorithm.
LoRMIkA tackles redundancy of features in rules and is able to generate simple rules.
Furthermore, we are able to search for the ``k'' number of best rules with respect to confidence, lift, leverage, coverage and support, which are the measures developed by the association rule mining community~\citep{Geng2006-ee} to measure if rules are predictive and/or interesting.

The state-of-the-art algorithms currently produce two types of rules which 1) explain the prediction towards the decision and 2) explain with counterfactual rules how to reverse the current decision of the global model. LoRMIkA generates two additional types of rules (four types of rules in total) that provide additional information and explanations for the prediction provided by the global model. Section~\ref{sec:fourtypes} explains the different types of rules in detail and illustrates and demonstrates how they can be used for explanations.

%We explain all four types of rules in the following example. 
%
%Let us assume a binary classification problem where the global model predicts $target=1$ with a probability of 70\% and $target=0$ with a probability of 30\%, accordingly. To explain this prediction of the global model LoRMIkA generates rules to answer the following questions.
%\begin{itemize}
%	\item What are the conditions that support the global model prediction to achieve the predicted class (target=1 and the potential conditions to achieve the probability of 70\%)?
%	\item What are the conditions that support the global model prediction to deviate from the predicted class (target=0 and the potential conditions to achieve the probability of 30\%)?
%	\item What are the hypothetical conditions to increase the predictive probability towards the predicted class (target=1 and the potential conditions to increase the probability of more than 70\%)?
%	\item What are the rules that could potentially reverse the prediction from predicted class to other classes (the potential conditions to reverse the target into ``0'')?
%\end{itemize}

We have tested and compared LoRMIkA with other state-of-the-art approaches on four different tabular classification datasets and conducted both qualitative and quantitative experiments to assess both interpretability and local accuracy, with competitive results.

In summary, the contributions of this paper are as follows.
\begin{itemize}
	\item We propose a novel approach based on class association rules to generate local rule-based explanations for the predictions of black-box machine learning models for classification problems.
	\item Our approach produces four types of rules which explain the prediction, namely: conditions that positively support the decision of the prediction, conditions that negatively support the decision prediction, conditions to increase the probability towards the decision, conditions that potentially reverse the decision.
	\item In our approach, rules can be searched for different optimization goals (i.e., coverage, lift, confidence etc.) to generate both predictive and/or interesting rules based on the user preference.
\end{itemize}

The remainder of this paper is structured as follows. Section~\ref{sec:assoc} revisits the most important results from the association rule mining community with respect to local explainers. Section~\ref{Sec:Approach} presents our approach. Section~\ref{Sec:results} discusses the experimental setup and results, and Section~\ref{Sec:conc} concludes the paper.

\section{Association Rule Mining}
\label{sec:assoc}

In this section, we revisit the major relevant findings in association rule mining research with respect to association rules being local models, predictive versus explanatory rules, and efficient ways for finding rules.

%\section{Local Interpretability and Association Rule Mining} \label{Sec:Local Interpretability and Association Rule Mining}

%In this section we discuss notations, definitions, and background related to the main topic of the paper along three aspects: local model-agnostic interpretability, rule-based explanations, and association rule mining.

\subsection{Notations and Definitions}
\label{Sec:Assocmining}

Association rule mining is a rule-based machine learning approach to find interesting combinations of input and output variables of frequent patterns, correlations, associations, or causal structures in large databases. It was firstly introduced by \citet{Agrawal1993-wa}. In the following, we define the concepts used in our work.
%
%Let $r = p \rightarrow q$ be a class association rule.

%
%Association rule mining, as first introduced by \citet{Agrawal1993-wa}, can be used to extract such rules from data. In the following, we define the concepts used in our work.

\begin{defn} \label{Def:AssociationRules}
Class Association Rule: A class association rule is a rule of the form $r = p \rightarrow q$. The antecedent or left-hand side (LHS) of a rule, $p$, is a boolean condition on feature values. The consequent or right-hand side (RHS) of a rule, $q$, is the target class of the decision variable. 
\end{defn}

\begin{defn} \label{Def:Support}
	Support: The proportion of instances that match p from the total amount of instances in the dataset.
	\begin{equation}
	\text{Support}(p) = \dfrac{|\text{instance} \in \text{Dataset, such that instance fulfills p}| }{\text{Total number of instances in the dataset}}
	\end{equation}
	
		\begin{equation}
	\text{Support}(p \rightarrow q) = \text{Support}(p \wedge q)
	\end{equation}
	
\end{defn}
\begin{defn} \label{Def:Coverage}
	Coverage:  The support of the antecedent of an association rule. 

	\begin{equation}
		\text{Coverage}(p \rightarrow q) = \text{Support}(p)
		\end{equation}
		
\end{defn}
\begin{defn} \label{Def:Strength}
	Confidence (i.e., Strength or Precision):  The percentage of instances in the dataset which contain the consequent and antecedent together over the number of instances which only contain the antecedent.
	\begin{equation}
	\text{Confidence}(p \rightarrow q) = \dfrac{\text{Support}(p \rightarrow q)}{\text{Support}(p)}
	\end{equation}
\end{defn}
\begin{defn} \label{Def:Lift}
	Lift:  Measures how often feature values in $q$ appear in instances that contain $p$, while controlling for the frequency for target value $q$. This is a measure for how surprising or interesting a rule is, i.e., how much it differs from a random association, in the sense that it represents a non-trivial correlation between antecedent and consequent~\citep{Webb2006-bi}. A lift equal to one implies that no associations can be found. A lift between zero and one means a negative association and a lift greater one means positive association.
    \begin{equation}	
	\text{Lift}(p \rightarrow q) = \dfrac{\text{Support}(p \rightarrow q)}{\text{Support}(p)\times \text{Support}(q)}
	\end{equation}
\end{defn}
\begin{defn} \label{Def:Leverage}
	Leverage: Calculates the observed frequency of $p$ and $q$ appearing together minus the frequency that would be expected if $p$ and $q$ were statistically independent. 
	Leverage can vary between $[-1,1]$. When the leverage is equal to zero, no associations can be found. When it is positive/negative, a positive/negative relationship of the antecedent and consequent can be determined. 
\begin{equation}	
	\text{Leverage}(p\rightarrow q) = \text{Support}(p\rightarrow q)- (\text{Support}(p)\times\text{Support}(q))
	\end{equation}
\end{defn}

\subsection{Advantages of using association rules as local explainers}

In this section we discuss the advantages of considering association rules as local explainers over other interpretable models such as decision trees or linear models.

\subsubsection{Rules versus feature importance of linear models}

The prediction of a linear model provides the target as a weighted sum of the feature inputs, and therewith can represent only linear relationships rather than non-linear relationships. Another drawback of linear models is that if there are input variables that are highly correlated, the estimated coefficients of the model can be high for either of the correlated features, which does not affect the predictive abilities of the model but leads to poor explanations.
%Moreover, decision rules are more interpretable and help to make decisions easily compared with the feature importance based representations produced by the linear models. 

%\textbf{TODO: I'm not sure if maybe also the rules model only linear relationships. The main drawback of linear models are that the coefficients are quite arbitrary for highly correlated features etc. Another drawback is that they also still build a global model w.r.t. the neighbourhood, but I think that's exactly what you discuss in the next section.}

\subsubsection{Association rules are local models}
A typical machine learning algorithm for prediction will have to
produce a single global model that is often the result of some form of model selection. 
In this process, interesting findings of the
dataset that may contain useful explanations may get lost and not be considered, as usually there will be not one single valid model, but different potentially equally valid ones.
\citet{Webb2011-cu} discussed the characteristics of association
rules being local models, as they consider only certain features and
only certain values of these features, thus only considering a
subspace of the feature space. 
Association rule mining aims to discover all such local models. For example, if there are two equally predictive rules, we aim to find both of these rules which help users when making decisions. 
Also, the globally optimal model may not be the optimal solution for a locally defined region in the subspace.
Association rule mining can find the optimal models in any specified region, which will be more efficient than a global model~\citep{Webb2011-cu,Martin2016-hb}.
 
%\textbf{TODO: I don't quite understand this. What is simple to explain about it?}
%Naturally, association rules are local models which are local in feature aspect rather than at the instance level.

\subsubsection{Rules that predict vs rules that explain}
\label{SubSec:Mining intersting rules}
%\textbf{TODO: I think the main thing that you want to say here is that prediction and explanation are different things and that something that predicts well doesn't necessarily explain well, and vice versa.}
Predictability of a rule can be measured using the confidence of the rule and the interestingness of a rule can be measured by the lift or the leverage. Furthermore, when the value of the confidence is high, the rule is said to be a predictive rule and when the value of the lift or the leverage is high, the rule is said to be an interesting rule. \citet{Novak2009-iv} discuss the differences between interpretability and predictability, by showing that the most predictive rules and the rules that explain best on a given dataset will be usually different. 
Using the example of a C4.5 decision tree for a predictive algorithm, they illustrate that redundant rules will be ignored, while
in descriptive algorithms, redundant rules should be considered. 
On the other hand, highly predictive rules may result from spurious correlations in the training data, if they represent only a small number of examples. 
Such rules will be filtered out by an adequate descriptive algorithm accordingly, while a predictive algorithm may be forced to take such rules into account for the sake of completeness of the predictions. 
Thus, though rules may not be predictive, they may still be of
interest to understand a dataset, and on the other hand, even if rules are highly predictive they may not be useful for explanations.

\subsubsection{Mining interesting rules efficiently}

The traditional approach for association rule mining is the Apriori
approach~\citep{Agrawal1993-wa} whose fundamental step is to find the
itemsets that occur most often, the so-called frequent itemsets. 
To select frequent itemsets, a minimum support needs to be given as a parameter.

Association rules are then determined within these frequent itemsets.
The frequent itemset association paradigm has a number of drawbacks,
such as its inability to uncover higher order associations, as these
are comparatively infrequent.
This is known as the so-called vodka and caviar problem~\citep{Cohen2001-qn}. 
Another limitation of the approach is that minimum support is not a reasonable threshold to regulate the number of associations, as it is impossible to select the number of associations in advance. 
So it boils down to a trial and error process, and therewith the minimum support constraint is not a faithful parameter to acquire interesting association rules~\citep{Webb2011-cu}.

\citet{Webb1995-cu} presents the Optimized Pruning for Unordered Search (OPUS) algorithm, that overcomes many of the shortcomings of other association rule mining techniques. It employs a statistically sound process of selecting the top k interesting rules, i.e the rules with the highest support, coverage, confidence, leverage, or lift. Furthermore, it rigorously controls the generation of spurious rules~\citep{Webb2011-cu} and has the ability to enable filter modes to adjust the
length of the rules to be discovered according to input parameters. When generating interesting rules (rules with a non-trivial correlation between antecedent and consequent), to overcome false positives OPUS uses a Fisher's exact hypothesis test~\citep{Webb2006-bi}.

\section{Our Approach} \label{Sec:Approach}
The main goal of our approach LoRMIkA is to explain an instance locally, with k-optimal class association rules. 
%
%  "Widely discussing" a nonsensical phrase

\subsection{Formal definition}

In local model-agnostic interpretability, we retain the global model as a  black box and explain the decision (or prediction) provided by the global model for each individual instance~\citep{Guidotti2018-cn}. Let $f$ be a given global black-box model, $x$ be an instance to be explained, and let $y$ be the prediction provided by the global model for $x$, i.e., $f(x) = y$. The aim is then to provide an explanation $e$ for the prediction $y$. Therefore, we use a local explainer model $e$ where $e=\xi(f,x)$, which mimics the behaviour of $f$ for $x$ using a process $\xi(.,.)$. The local interpretability model considers the neighbourhood of the instance $x$ in the global model $f$ to provide the explanation $e \in E$ that belongs to a human-interpretable domain $E$. 
In our case of rule-based explanations, $E$ is a domain of rules, i.e., an explanation $e = \{r_1, r_2, ...\}$ provides a set of class association rules $r$ which explain the prediction $y$ from $f$ for the instance $x$.

%An explanation $e \in E$ is based on $f'$, if $e=\psi(f',x)$ for some explanation logic $\psi(.,.)$ which reasons over $f'$ and $x$. \textbf{TODO: What is psi? psi is the explanation logic. for  example by just applying association rule mining to the neighbourhood, it will give somerules. we have an another logic to get 4 kinds of rules. that logic is $\psi$. Either explain this better or get rid of psi and replace $f'$ with $e$.} 

\subsection{Overview of the procedure}

\begin{figure*}[t]
	\centering
	\includegraphics[width=1.01\textwidth]{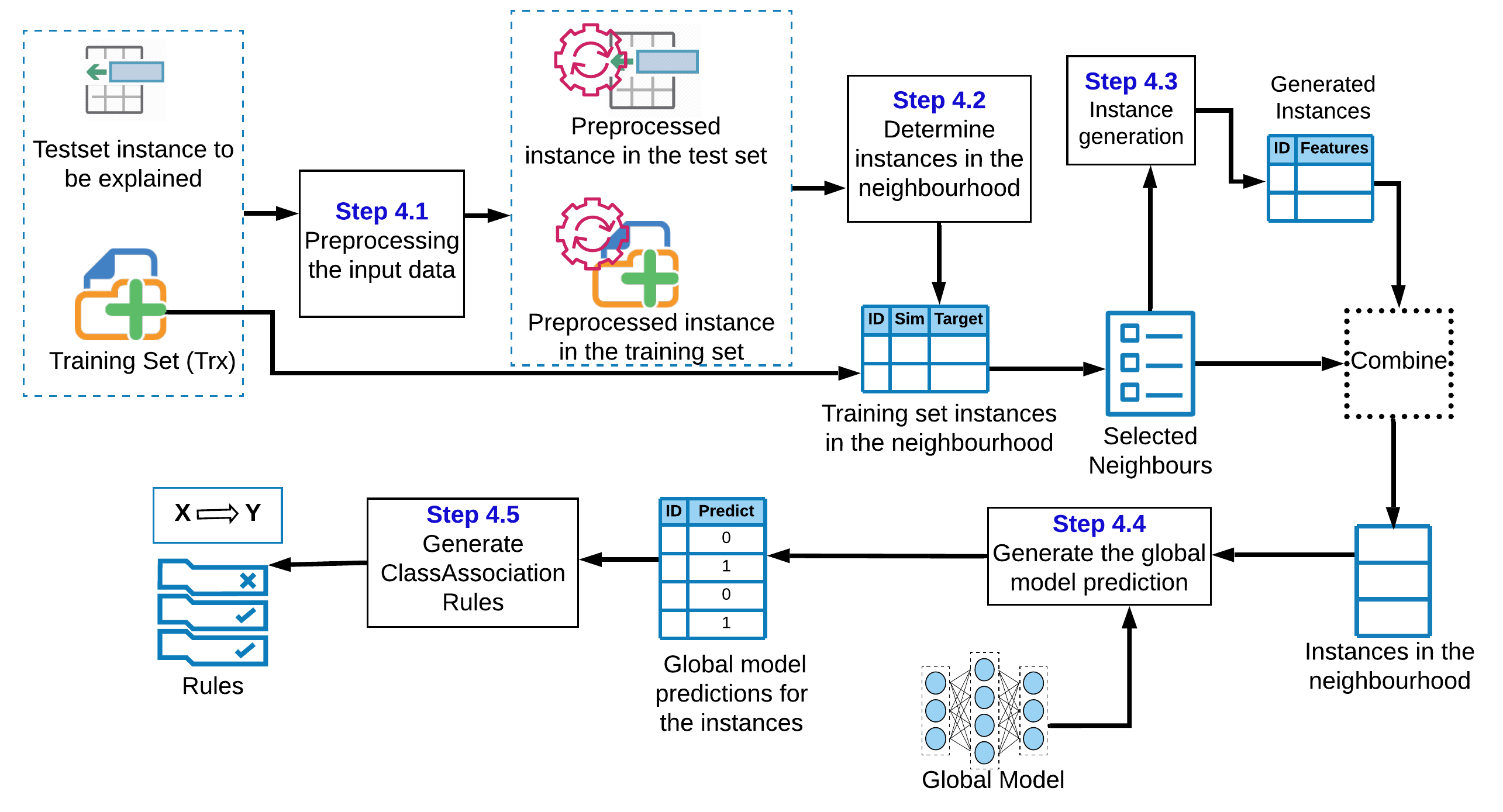}
	\caption{An overview diagram of LoRMIkA to generate local rule-based explanations for the global model predictions.}
	\label{Fig:framework}
\end{figure*}

Broadly, we first generate a neighbourhood that has similar behaviour to the instance to be explained. For that, first, we preprocess the input data (i.e., training data and instances to be explained). Then, we select the neighbourhood of the instance to be explained with the help of a distance measure. Then, we generate new instances in the neighbourhood with mutation and crossover/interpolation techniques. After that, we generate global model predictions for the whole set of instances in the neighbourhood (i.e., generated  and selected training instances) to learn the behaviour of the global model. Finally, we perform association rule mining using the OPUS search~\citep{Webb1995-cu} algorithm with k-optimal~\citep{Webb2011-cu} associations for the combined set of instances and their global model predictions to produce class association rules as the explanations of the prediction. The resulting rules are then categorized into four categories based on the feature values and the predicted value of the global model of the instance to be explained.
An overview diagram of LoRMIkA is shown in Figure~\ref{Fig:framework}. 
Our approach is furthermore summarized in Algorithm~\ref{Alg:WholeMethod} and the most important parts are discussed in the following.

\begin{figure*}[t]
	\centering
	\includegraphics[width=\textwidth]{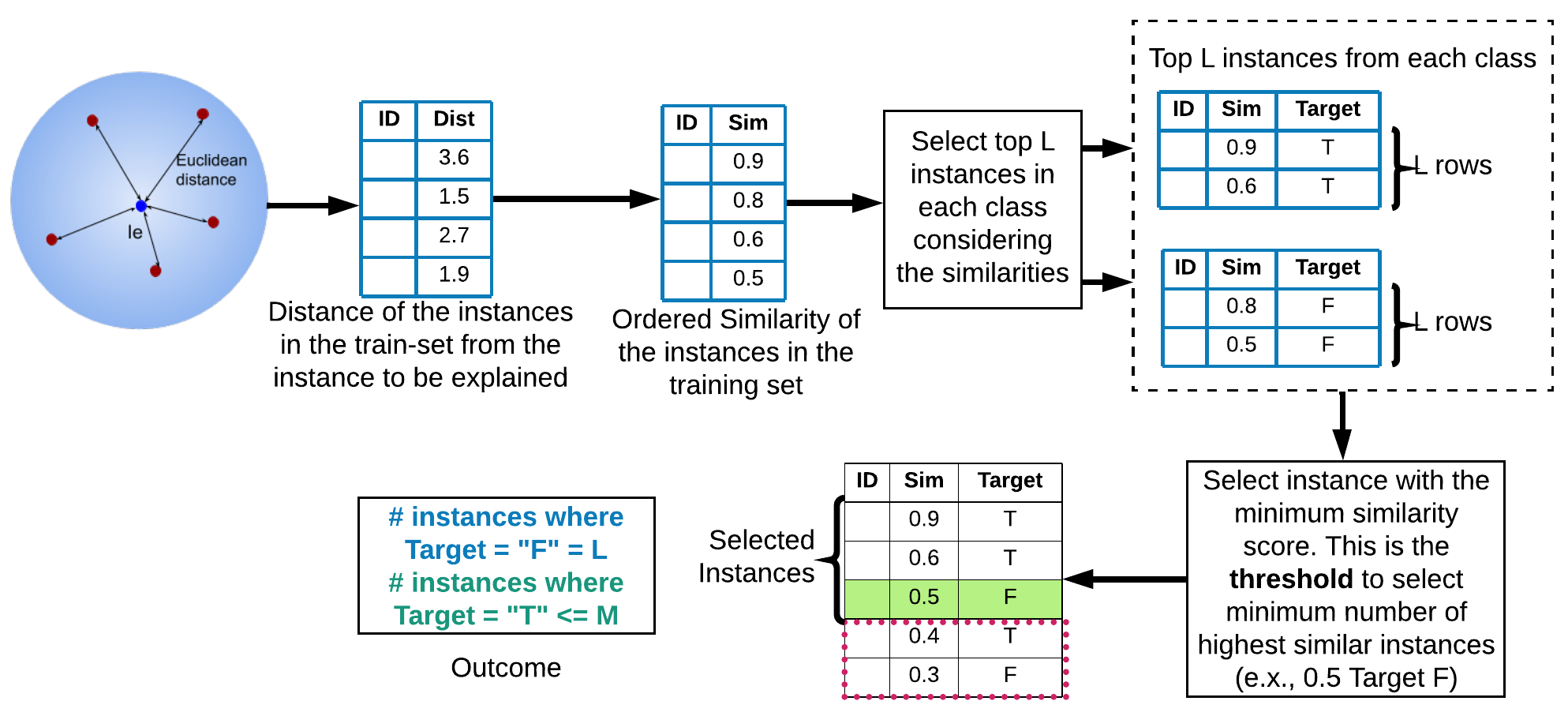}
	\caption{An approach to determine instances in the neighbourhood}
	\label{Fig:selected_ins}
\end{figure*}

%\begin{comment}
%%%%%%%% FIGURE %%%%%%%%% 
%\begin{figure}[t]
%\includegraphics[width=\linewidth]{local_explainer}
%\caption{General process of the local explainer model. Assume that the red cross (x) is the instance to be explained. Blue and purple crosses are the training data (background dataset) from two classes. Here, we have selected the instances in the background dataset which are in the neighbourhood of x. Purple and blue colour dots are the newly generated instances by applying cross-over and mutation techniques to the neighbourhood instances.}
%\label{Fig:local_explainer}
%\end{figure}
%%%%%%%% FIGURE %%%%%%%%%
%\end{comment}

\begin{algorithm}[t]
	\caption{Explain Black-box}\label{Alg:WholeMethod}
	\DontPrintSemicolon
	\SetKwInOut{Input}{Input}
	\SetKwInOut{Output}{Output}
	\Input{$Tr_x - \text{Training instances without target}$}
	\nonl\myinput{$Tr_y - \text{Target of training instances}$}
	\nonl\myinput{$I_e - \text{Instances need to be explained}$}
	\nonl\myinput{$M - \text{Global black-box model}$}
	\nonl\myinput{$F_c - \text{Categorical features}$}
	\nonl\myinput{$F_n - \text{Numerical features}$}
	\nonl\myinput{$N_\mathrm{GenInst} - \text{\#new instances to be generated}$}
	\Output{Rule-based explanations}
	\BlankLine
	\SetAlgoLined
	%	$TxtNam \gets \text{Make.namFile}(\text{colNames}(E,Tr_y),C,N)$\;
	$Tr_{px},I'_e \gets \text{PreProcessingInputData}(Tr_x,I_e)$\;
	\For{$i_e \in I'_e$}{
		$SelecInst \gets \text{SelectInstFrmNeighbourhood}(Tr_{px},i_e,Tr_x)$\;
		$GenInst \gets \text{GenInstFrmNeighbourhood}(SelecInst,F_c,F_n,N_\mathrm{GenInst},i_e)$\;
		$CombInst \gets SelecInst \cup GenInst $\;
		$Pred \gets \text{GetPredictFrmGlobalModel}(CombInst,M)$\;
		$GenRules \gets \text{MOGenRules}(CombInst,Pred)$\;
		\Return $GenRules$			
	}
\end{algorithm}

\subsection{Preprocessing the input data}
%In this step, we pre-process input data from the training set and the
%instances that are to be explained to find the nearest
%neighbours to the instance to be explained in a more efficient
%procedure. 
This step corresponds to the $\text{PreProcessingInputData}(Tr_x,I_e)$ function in Algorithm~\ref{Alg:WholeMethod}.
We preprocess the training data by applying Z-score normalisation for each numerical feature and one-hot encoding for each categorical feature, since the distribution of each studied feature varies in range. Note that depending on the distribution of the features, other normalisation methods such as min-max scaling may yield better results, however Z-score normalisation is arguably a good default choice for the generic case.
\begin{algorithm}
	\caption{$\text{SelectInstFrmNeighbourhood}(Tr_{px},i_e,Tr_{x})$} \label{Alg:SelectInstFrmNeighbourhood}
	\DontPrintSemicolon
	\SetKwInOut{Output}{Output}
	\KwIn{$Tr_{px} - \text{Pre-processed data}$}
	\KwIn{$i_e - \text{Instance to be explained}$}
	\KwIn{$Tr_{x} - \text{Training instances before pre\_processed}$}
	\Output{Instances in the neighbourhood of the selected instance that need to be explained.}
	\BlankLine
	\SetAlgoLined
	%	$H \gets \text{Maximum amount of neighbours from the training data }$\;
	$L \gets \text{Minimum number of neighbours from each class}$\;
	$M \gets \text{Maximum number of neighbours from each class}$\;
	%	$M \gets 5 \times L$\;
	%	$S_{ct} \gets \text{Cut point of the similarity score}$\;
	$w \gets \text{Kernel width of the similarity score}$\;
	$DistTrSet \gets \text{GetEuclideanDistance}(Tr_{px},i_e)$\;
	$SimTrSet \gets \text{CnvrToSimilarityScore}(DistTrSet,w)$\;
	$SimTrSetGroupScores \gets \text{GroupByClzSimilarityScore}(SimTrSet)$\;
	$SimTrSetGroupDesOrder \gets \text{OrderByClzSimilarityScore}(SimTrSetGroupScores)$\;
	$MinClzSimScoreLst \gets {[ ]}$\;
	\For{$e_{clzsim} \in SimTrSetGroupDesOrder$}{
		
		$MinClzSimScore \gets \text{SelectMinSimScoreTopL}(e_{clzsim})$\;
		$MinClzSimScoreLst.insert(MinClzSimScore)$\;
	}
	$S_{ct} \gets min(MinClzSimScoreLst)$ \tcc{Cut point of the similarity score}
	$InstGreaterSct \gets GetInstGreaterSct(S_{ct}, SimTrSet,Tr_{x})$\;
	$InstGreaterSctGroupLst \gets GroupByClz(InstGreaterSct)$\;
	$SelectInst \gets {[ ]}$\;
	\For{$e_{InstGtrSctGrp} \in InstGreaterSctGroupLst$}{
		\uIf{$e_{InstGtrSctGrp}.size() > M$ }
		{$SelectInst.insert(e_{InstGtrSctGrp}.choice(Inst_{M}))$} 
		\Else{
			$SelectInst.insert(e_{InstGtrSctGrp})$\;
		}
	}

	\Return $SelectInst$
\end{algorithm}

\subsection{Determine the instances in the neighbourhood}

In this step, we find the training set instances which are in the neighbourhood of the instance to be explained since we assume that the behaviour and the properties of the instances in the neighbourhood are resembling the instance to be explained.  
The process of instance selection within the given neighbourhood is
outlined in Algorithm~\ref{Alg:SelectInstFrmNeighbourhood} and Figure~\ref{Fig:selected_ins}. 
There are many ways for selection, e.g., \citet{Yu2018-di} use the Kullback-Leibler (KL) divergence. The process in our approach is similar to the approach proposed by~\citet{Yu2018-ou} when grouping users according to their similarities in social behaviour. 
A detailed explanation of the instance selection process is provided in the following. 

%\textbf{Step 1:}
First, the distance between each pre-processed instance in the training set (background set) and the instance that needs to be explained is calculated. 
We use Euclidean distance as the distance measure to calculate the distance between each pre-processed instance in the training set and the instance that needs to be explained, as it is preferrable to other measures in the interpretability context~\citep{ribeiro2016should, DBLP:journals/corr/abs-1805-10820}. Similarly,~\citet{ribeiro2016should} used other distance measures such as cosine distance for text, and L2 distance for images.
%Moreover,~\citep{Yu2019-co} considered Euclidean distance in order to classify the positive and the negative samples of the images.
%Since our proposed approach is suitable only for tabular classification datasets, the features of the dataset can be directly used as concepts for explanations.
As we use the features directly as concepts for explanations, we effectively limit our approach to situations where single features can provide meaningful explanations. 

We then convert the distance to an exponential similarity score to make the distance more linear and to compare the similarity of the instances in the training set with the explained instance. The similarity is defined as follows:

\begin{equation} \label{Eq:Exponential_similarity}
K(x,x') = \exp\bigg(-\dfrac{{x-x'}^2}{2w^2}\bigg).
\end{equation}

In Equation~\ref{Eq:Exponential_similarity}, $K(x,x')$ is the similarity between two instances $x$ and $x'$, and $w$ is the kernel width (see also Algorithm~\ref{Alg:SelectInstFrmNeighbourhood}), a tuning parameter. When $w$ is high, $K(x,x')$ will be close to $1$ for any $x,x'$. When $w$ is low, $K(x,x')$ will be close to $0$. Following \citet{ribeiro2016should}, we set $w$ to 0.75 times the number of features.

%This step is equal to $\text{CnvrToSimilarityScore}(DistTrSet,w)$ function in Algorithm~\ref{Alg:SelectInstFrmNeighbourhood}.

After that, we group the similarity scores according to the target class and we sort them in descending order. Then we select the instances with top L (i.e., minimum number of neighbours from each class) similarities in each class. 
From the selected top L instances over all the target classes, the similarity score of the instance with the lowest similarity is used as the threshold to select the minimum number of most similar instances. 
This threshold value is $S_ct$ (i.e., cut point of the similarity score) in Algorithm~\ref{Alg:SelectInstFrmNeighbourhood}. Further, considering the whole training set we select all the instances whose similarity score is more than $S_ct$ for further processing. 
%This step resembles line number 6-14 in Algorithm~\ref{Alg:SelectInstFrmNeighbourhood}.

In order to overcome the class imbalance problem, we check whether the number of instances from each class is less than or equal to M (i.e., the maximum number of neighbours from each class). 
Then, we sample M number of instances from the classes where the number of instances is greater than M, to select as a set of neighbours of the instance to be explained.  
Further, the instances of the classes where the number of instances is less than or equal to M are also considered as the neighbours of the instance to be explained. 
The output of Algorithm~\ref{Alg:SelectInstFrmNeighbourhood} (i.e., $SelectInst$) contains the most similar instances of the training set to the instance to be explained. 
%This step resembles to the step number 16-24 in Algorithm~\ref{Alg:SelectInstFrmNeighbourhood}. 

\subsection{Instance Generation}

%The instance generation procedure is outlined in Algorithm \ref{Alg:GenInstFrmNeighbourhood}.
%
The amount of instances chosen from the original training set may not be enough to adequately characterise the neighbourhood of the instance in consideration.
Therefore in our approach, we generate new instances using the neighbourhood instances of the training set based on crossover (i.e., interpolation) and mutation techniques, while ensuring that the majority of the new instances are inside the neighbourhood. 
The instance generation procedure is outlined in Algorithm \ref{Alg:GenInstFrmNeighbourhood}.
For the instance generation we perform the following steps. 

After determining the instances in the neighbourhood, in this step we use the output of Algorithm~\ref{Alg:SelectInstFrmNeighbourhood} to generate more instances within the selected neighbourhood. We generate 50\% of the total instances using the crossover technique defined in Equation~\ref{Eq:crossover_Equation}. It ensures that all the generated instances are within the neighbourhood.

\begin{equation} \label{Eq:crossover_Equation}
I_{\mathrm{crossover}} =x + (y-x)*\alpha
\end{equation}

Here, $x$ and $y$ are randomly selected instances from the training set in the neighbourhood of the instance to be explained, and $\alpha$ is a randomly generated number between $0$ and $1$.

Afterwards, we generate values for the categorical features for the newly generated instances. 
Here, we assume that the generated instance behaves similar to the instance to be explained if the generated instance values are similar to the values of the most similar parent instance to the instance to be explained.
Having that assumption, we select the most similar parent instance (i.e., $x$ or $y$ ) to the instance to be explained. The categorical values for the newly generated instance are equal to the categorical values of the most similar instance. 

Then, we use a mutation technique from \citet{Storn1997-wv}, defined in Equation~\ref{Eq:Mutation_Equation}, to generate the rest of the newly generated instances.

\begin{equation} \label{Eq:Mutation_Equation}
I_{\mathrm{mutation}} =x + (y-z)*\sigma
\end{equation}

Here, $x,y,$ and $z$ are three different instances of the neighbourhood of the instance to be explained, and $\sigma$ is a randomly generated number between $0.5$ and $1$. 
Similarly, when setting the values for the categorical features of the newly generated instances, we select the most similar instance from $x$, $y$, and $z$ to the instance to be explained and we set the values of the most similar instance to the newly generated instance.

\begin{algorithm}[!t]
	\caption{$\text{GenInstFrmNeighbourhood}(SelecInst,F_c,F_n,N_\mathrm{GenInst},i_e)$}\label{Alg:GenInstFrmNeighbourhood}
	\DontPrintSemicolon
	\SetKwInOut{Output}{Output}
	\KwIn{$SelecInst - \text{Output of the Algorithm \ref{Alg:SelectInstFrmNeighbourhood} : neighbours in the training set}$}
	\nonl\myinput{$F_c - \text{Categorical Features}$}
	\nonl\myinput{$F_n - \text{Numerical Features}$}
	\nonl\myinput{$i_e - \text{Instance to be explained}$}
	\Output{$GenInst - \text{Generated instances}$}
	\BlankLine
	\SetAlgoLined
	%	$t \gets \text{threshold}$\;
	$\alpha \gets \text{uniform($0,1$)}$\;
	$\sigma \gets \text{uniform($0.5,1$)}$\;
	$x,y \gets \text{RandomSelecTwoInst($SelecInst$)}$\;
	$I_\mathrm{crossover} \gets x + (y-x) * \alpha$\;
	$x,y,z \gets \text{RandomSelecThreeInst($SelecInst$)}$\;
	$I_\mathrm{mutation} \gets x + (y - z) * \sigma$\;
	\tcc{Get closest neighbours to the explained instance in each $I_\mathrm{crossover}$ and $I_\mathrm{mutation}$ case} 
	$closeInst_\mathrm{crossover} \gets \text{GetClosestInstance$(i_e,[x,y])$}$\;
	$closeInst_\mathrm{mutation} \gets \text{GetClosestInstance$(i_e,[x,y,z])$}$\;
	\begin{comment}
	\uIf{$dist(x,e) \leqslant dist(x',e)$}{
	$closeIns \gets x$ \;
	}
	\Else{
	$closeIns \gets x'$ \;
	}
	\end{comment}
	\ForEach{$f_c \in F_c$}{$I_\mathrm{crossover}(f_c) \gets closeInst_\mathrm{crossover}(f_c)  $\;
		$I_\mathrm{mutation}(f_c) \gets closeInst_\mathrm{mutation}(f_c)  $}\;
	\Return $I_\mathrm{crossover} , I_\mathrm{mutation}$
\end{algorithm}

\subsection{Generate global model predictions}

This step corresponds to $\mathrm{GetPredictFrmGlobalModel}(ComnInst,M)$ in Algorithm~\ref{Alg:WholeMethod}.
The assumption behind the generation of global model predictions is that the logic to produce the global model prediction of the instance to be explained is similar to the logics of producing the global model predictions of its neighbourhood instances. 
Here, we obtain the global model predictions of the newly generated instances and the neighbours taken from the training set (i.e., $SelecInst \cup GenInst$) to obtain the global model behaviour for this combined dataset. 
The intuition in this step is to generate a new training dataset in the neighbourhood of the instance to be explained to fit the local explainer (i.e., association rule mining algorithm) in the locally defined neighbourhood of the instance to be explained. This newly generated training set has the global model predictions as the target values.

\subsection{Generate class association rule-based explanations}
\label{sec:fourtypes}

This step relates to the $\mathrm{MOGenRules}(CombInst,Pred)$ function in Algorithm~\ref{Alg:WholeMethod}.
We generate k optimal class association rules using the OPUS search algorithm~\citep{Webb2011-cu} with regards to different objectives such as support, coverage, confidence, lift and leverage. Here, k is the maximum number of rules to be generated for each objective. In our approach (LoRMIkA) we provide both predictive rules (i.e., rules with high confidence) and interesting rules (i.e., rules with high lift). An implementation of OPUS search to generate association rules is available in the BigML platform~\citep{donaldson2012package}.

The reason for using the OPUS algorithm for rule extraction rather than algorithms like CPAR~\citep{Yin2003-fy}, RIPPER~\citep{Cohen1995-mp}, FOIL~\citep{Quinlan1993-np}, or  RCAR~\citep{Azmi2019-rg} is that those algorithms focus on finding minimal rule sets which lead to accurate predictions (i.e., high confidence).
As discussed in Section~\ref{SubSec:Mining intersting rules} the main drawback of focusing on the most predictive rules is the inability to uncover higher-order associations, which are relatively infrequent. The OPUS algorithm is a statistically sound algorithm which captures infrequent higher order associations (interesting rules) while ignoring false positives. As stated before, we argue that the explanations provided by the local explainers should provide good explanations and not necessarily good predictions.

\begin{figure*}[t]
	\centering
	\includegraphics[width=0.7\textwidth]{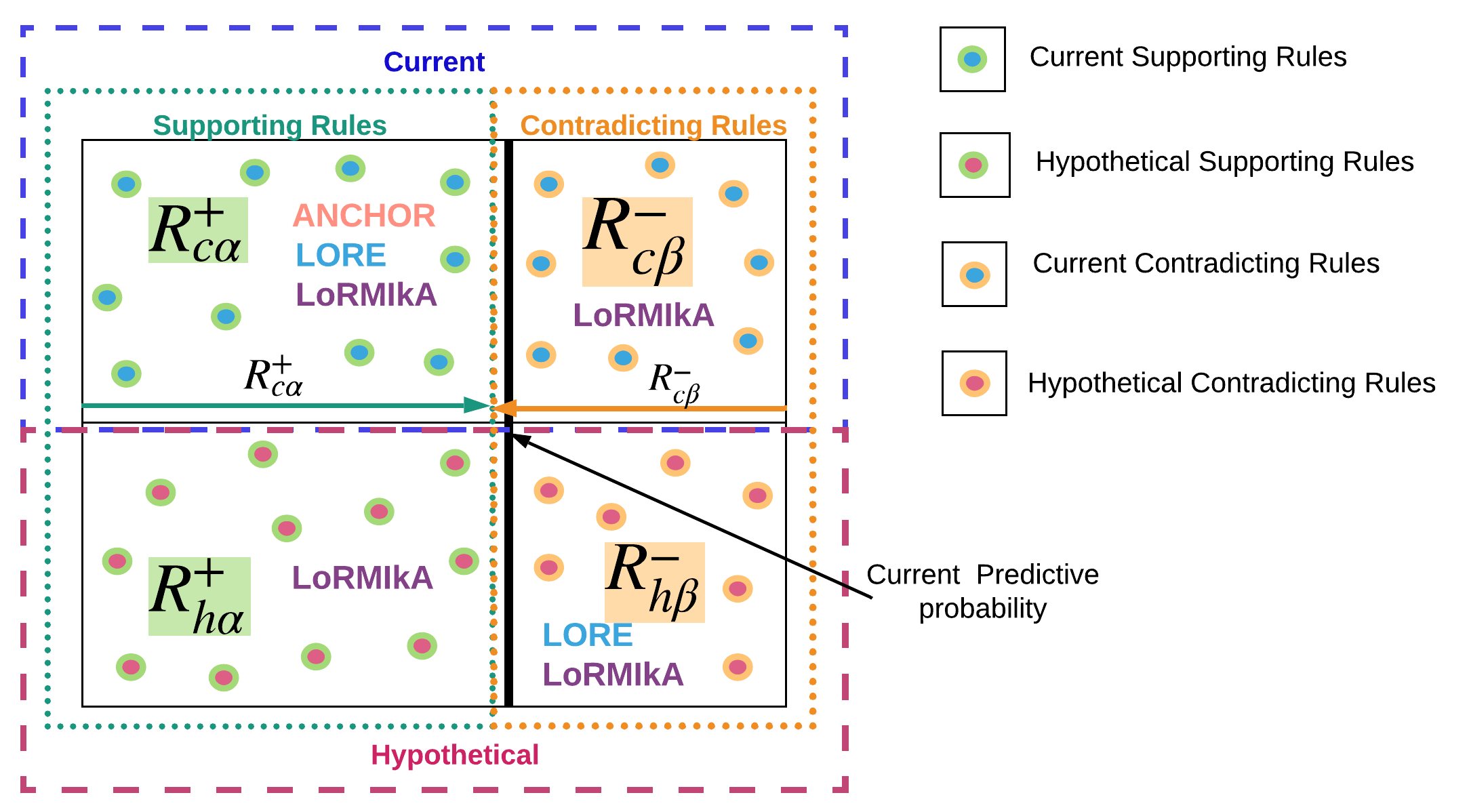}
	\caption{Diagram that illustrates the four types of rules generated by LoRMIKA.}
	\label{Fig:rules}
\end{figure*}

The inputs to the association rule generating OPUS algorithm are the instances in the neighbourhood of the instance that needs to be explained. Both instances that were selected from the training set and newly generated instances are considered. 
LoRMIkA provides flexibility to search for the k-optimal rules with the highest support, coverage, confidence, lift, or leverage. This optimisation criterion is a parameter of the method, and thus this choice can be made by the user, and both predictive and interesting rules can be generated.
Also, as there is in general the risk of over-estimating the confidence when the coverage is low, the OPUS algorithm uses an \textit{m-estimate}~\citep{Dzeroski1993-jl} to adjust the values of confidence and lift. 
This feature helps to avoid overfitting by discovering spurious highly predictive rules. 

We classify the set of rules obtained from the OPUS algorithm into four categories with regards to a contingency table of the LHS and RHS of the
association rules. 
The LHS of the rule is true if the antecedent of the rule agrees with the feature values of the instance to be explained. The RHS of the rule is true if the consequent of the rule agrees with the prediction provided by the global model of the instance to be explained. 

Let us consider an example in a cancer risk prediction scenario, where for a patient $x$ the global model predicts this patient as a positive case ($y=1$), i.e., the patient is at risk of developing a cancer. To explain this prediction LoRMIkA produces four types of rules as follows (also see Figure~\ref{Fig:rules}). 
\begin{enumerate}[1.]

\item{\textbf{Current Supporting rules ($\Re_{c\alpha}^+$ )}: The rules that support the prediction of the global model.
\\\indent \emph{Definition}: if LHS = true, then RHS = true.
\\\indent \emph{Interpretation}: Conditions of $x$ that support and explain why the patient is predicted as a risk case ($y=1$) by the global model. 
\\\indent \emph{Benefits}: Observing the current supporting rules, the doctors can identify the conditions that lead to classify this patient as at risk by the global model. These rules indicate the features that this patient has that lead to a positive classification. 
\\\indent \emph{Approaches}: Anchor, LORE and LoRMIkA provide current supporting rules. }

\item{\textbf{Current Contradicting Rules ($\Re_{c\beta}^-$ ):} The rules that contradict the prediction of the global model.
\\\indent \emph{Definition}: if LHS = true, then RHS = false.
\\\indent \emph{Interpretation}: Conditions of $x$ that contradict the current prediction ($y=1$) and would support its prediction as being negative ($y=0$). These rules contradict the current prediction ($y=1$).
\\\indent \emph{Benefits}: Observing the current contradicting rules indicates doctors the features that the current patient has that would lead to a negative prediction, i.e., that contradict the current prediction. In our example, these factors contribute to a prediction of not being at risk of cancer, and therewith are factors that the doctors should try not to change.
\\\indent \emph{Approaches}: Only LoRMIkA provides current contradicting rules.}

\item{\textbf{Hypothetically Supporting Rules ($\Re^{+}_{h\alpha}$) :} The rules that would increase the probability of the prediction of the global model.
\\\indent \emph{Definition}: if LHS = false, then RHS = true.
\\\indent \emph{Interpretation}: Determine the conditions that are currently not satisfied by $x$ (they are hypothetical) that would further support the current prediction of $x$ being a risk case. 
\\\indent \emph{Benefits}: If features of the patient changed towards the conditions in the rules, this would further increase the risk of $x$ to be positive. Thus, changes along the lines of these rules should be avoided in our example.
\\\indent \emph{Approaches}: Only LoRMIkA provides hypothetically supporting rules.}

\item{\textbf{Hypothetically Contradicting Rules (Counterfactual Rules, $\Re^{-}_{h\beta}$ )}: The rules that may invert the prediction of the global model.
\\\indent \emph{Definition}: if LHS = false, then RHS = false.
\\\indent \emph{Interpretation}: The conditions that are currently not satisfied by the features of $x$ (they are hypothetical) that would contradict the current prediction of $x$ as positive, i.e. they would contribute towards inverting the prediction to being negative ($y=0$, \emph{not} being at risk).
\\\indent \emph{Benefits}: The counterfactual rules can help doctors to understand in which ways the characteristics of the patient needed to change to make the patient more likely to be classified as a negative patient.
\\\indent \emph{Approaches}: LORE and LoRMIkA provide hypothetically contradicting rules. }
\end{enumerate}

With these four kinds of rules, our approach is able to provide a complete picture of rules that explain the local neighbourhood of an instance and give the practitioner all the information for informed decision making.

A Python implementation of LoRMIKA is available from GitHub\footnote{https://github.com/DiliSR/LoRMIkA}. It can be used together with an implementation of the OPUS class association rule mining algorithm, such as the one from BigML~\citep{donaldson2012package}.
%\footnote{https://github.com/papersubmission123456/icse2020-explainable}.

\section{Experiments}\label{Sec:results}

In this section, we compare LoRMIkA with other state-of-the-art rule-based algorithms in both qualitative and quantitative aspects.

\subsection{Overview of the datasets}

We use the evaluation framework from \citet{DBLP:journals/corr/abs-1805-10820} for our experiments, which employs three real-world classification datasets Adult~\citep{Ronny1996-be}, COMPAS~\citep{Jeff2017-qu}, German~\citep{Hofmann1994-on}. Furthermore, we use the Covertype~\citep{Dua:2019} dataset to evaluate our method on larger datasets. The datasets have both categorical and continuous features. Continuous features are discretized in our approach. In the Adult, COMPAS, and German datasets, each instance of a dataset represents a record that belongs to an individual. In the Covertype dataset, each instance represents a $30m \times 30m$ patch of forest that is classified as one of seven cover types. 

%\textbf{TODO: I don't quite understand how the covtype dataset looks like.}

The \textbf{Adult} dataset from the UCI Machine Learning Repository contains data about income levels in relation to demographic features. It contains 48,842 instances in total. The income, which is
the target column of the dataset divides the whole dataset into two
classes: \enquote{$\leq50K$} and \enquote{$\geq50K$}.

The \textbf{COMPAS} dataset from ProPublica includes the features used
by the COMPAS algorithm to assess a criminal defendant’s likelihood to
re-offend (Low, Medium and High). This dataset consists of over 10,000
instances and we have divided it into the two classes ``Medium-Low'' and ``High'' risk.

The \textbf{German} dataset from the UCI Machine Learning Repository contains 1,000 instances and classifies persons as \enquote{good} or \enquote{bad} creditors.

The \textbf{Covertype} dataset from the UCI Machine Learning Repository contains 581,012 instances. 
It is a multi-class classification dataset with seven classes. 
As multi-class classification is not the main focus of our paper, we convert it into a binary classification problem as follows.
We consider the class which has the highest number of instances to be one class (``type = 2'') and merge all the other classes into the second class (``type = 1''), making it a one-vs-all classification.

%\textbf{TODO: Did you binarise the covertype dataset?}

%\begin{comment}
%	In the German dataset from UCI Machine Learning Repository each person of the 1,000 entries is classified as a “good” or “bad” creditor according to attributes like age, sex, checking_account, credit_amount, duration, purpose, etc.
%\end{comment}
\subsection{Generate global model predictions}

To generate predictions from global models, we perform the following steps.
First, we select a set of machine learning models as global models to make predictions in our experiments.
For the experiments, we consider five algorithms as the global
black-box models. They are support vector machine (SVM), random forest (RF), logistic regression (LR), decision trees (DT) and
multi-layer neural networks with `LBFGS' solver (NN). 
We use these models in their implementations from the scikit-learn library~\citep{pedregosa2011scikit}, with their default parameters. The only parameter that we change from the default parameters is in the RF algorithm, where we set the number of trees to 100, to keep the hyper-parameters of the global models of our experimental setup consistent with the experimental setup of LORE~\citep{DBLP:journals/corr/abs-1805-10820}.
Some of these models, for instance DT, can be considered interpretable globally. However, we argue that it is still worthwhile to generate local explanations, as they are targetted at the particular instance and can therewith be more relevant.
We do not consider the SVM model for the Covertype dataset in our experiments, as it is computationally too expensive and we did not obtain results after 10 days of running time. 

After preprocessing the datasets we impute missing values in both continuous and categorical variables of the dataset using mean and mode imputation, respectively. Moreover, we discretise continuous attributes into three sub-ranges, where the size of the sub-ranges is chosen for them to contain an approximately equal number of instances.
Finally, to select the training set and testing set each dataset is randomly split
using a 4:1 ratio.

\subsection{Select rule-based model agnostic explainers for baseline comparison}

For the baseline comparison, we use Anchor~\citep{Ribeiro2018-mc} and LORE~\citep{DBLP:journals/corr/abs-1805-10820} as two state-of-the-art local rule-based model-agnostic explainers.

\paragraph{Anchor} The main goal of Anchor is to provide \emph{if-then} rules, which are called \emph{anchors}. It provides rules with high confidence, where it guarantees that changing the values of the features which are not included in the rule will not affect the final prediction of the global model. Also, Anchor selects the rules with a minimum confidence of 95\%. If multiple rules are generated with the same confidence, then it selects the rule with the highest coverage from those.
As the output, Anchor provides a single current supporting rule which explains the prediction of the global model (i.e., $\Re_{c\alpha}^+$).

\paragraph{LORE} First, LORE generates new instances in the neighbourhood of the instance to be explained using a genetic algorithm. 
To learn the behaviour and the properties of the global model, it obtains the global model predictions of the generated instances.
Finally, LORE obtains rule-based explanations after building a decision tree on the generated instances. 
As the output, LORE provides a single current supporting rule which explains the prediction of the global model (i.e., $\Re_{c\alpha}^+$) and rules which potentially help to invert the global model predictions which are called counterfactual rules (i.e., $\Re_{h\beta}^-$).

\subsection{Parameter setup in LoRMIkA}

In preliminary experiments, we determined that the parameters of our method are fairly robust, and consequently we use a common set of parameters in all three datasets as follows. 
From Algorithm~\ref{Alg:SelectInstFrmNeighbourhood}, $L = 40$, $w = 0.75 \times \sqrt{\text{no. of features}}$. We choose the target proportion of minority class instances to the majority class instance as 1:5. All runs are executed 10 times and averages are reported here, for the stability of the results.

%\begin{comment}
%%%%%%%% FIGURE %%%%%%%%% 
%\begin{figure}[t]
%\centering
%\includegraphics[width=0.5\textwidth]{instance_exp}
%\caption{Example instance to be explained}
%\label{Fig:instance_exp}
%\end{figure}
%%%%%%%% FIGURE %%%%%%%%%
%
%
%%%%%%%% FIGURE %%%%%%%%% 
%\begin{figure}[t]
%\centering
%\includegraphics[width=0.5\textwidth]{lormika_new}
%\caption{Example explanation for an instance of the COMPAS dataset from LoRMIkA}
%\label{Fig:alg_output}
%\end{figure}
%%%%%%%% FIGURE %%%%%%%%%
%
%\end{comment}
%\begin{comment}
%%%%%%%% FIGURE %%%%%%%%% 
%\begin{figure}[t]
%\centering
%\includegraphics[width=0.6\textwidth]{anchor_new}
%\caption{Example output explanation of the Anchor algorithm (same instance used for all algorithms in the example). No counterfactual rules are obtained.}
%\label{Fig:anchor_output}
%\end{figure}
%%%%%%%% FIGURE %%%%%%%%%
%
%
%%%%%%%% FIGURE %%%%%%%%% 
%\begin{figure}[t]
%\centering
%\includegraphics[width=0.5\textwidth]{lore_new}
%\caption{Example output explanation of the LORE algorithm with a rule for the decision and a counterfactual rule (same instance used for all algorithms in the example).}
%\label{Fig:lore_output}
%\end{figure}
%%%%%%%% FIGURE %%%%%%%%%
%
%\end{comment}

\subsection{Qualitative analysis}

In this section, we illustrate with an example the performance of Anchor, LORE, and LoRMIkA. We consider a prediction for an instance of the COMPAS dataset. 
We use the random forest classifier that classifies the instance in consideration as having a 65\% probability for the prisoner of having a ``High'' level of likelihood to re-offend. The feature values of the instance are as follows:

\begin{center}
\label{Fig:Instance}
		
	\begin{tabular}{ ccc } 
		%	\toprule
		\multicolumn{3}{c}{	{\text{Prediction}} = {High (65\% probability)}}\\
		\midrule
		%\hline
		age = 27 & age\_cat = 25 - 45 & sex = Male \\ 
		race = African-American & priors\_count = 4 & days\_b\_screening\_arrest = 1 \\ 
		c\_charge\_degree = F & is\_recid = 1 & is\_violet\_recid = 1 \\ 
		two\_year\_recid = 1 & length\_of\_stay = 50 &  
		%\hline
		%	\bottomrule
	\end{tabular}

\end{center}

In the following, we discuss the possibilities of releasing the prisoner and its implications, using the local explainer models. % that need to be taken and should not be taken by the relevant authorities  before releasing this prisoner to society. 

\paragraph{What are the current conditions that support the prediction as having a High chance to re-offend?} This question can be answered using the current supporting rules ($\Re^+_{c\alpha}$). The current supporting rules of LoRMIkA are as follows.
 
\begin{align}
\Re^+_{{c\alpha\_}{{\text{LoRMIkA}}}} &=  \{( 26 \leq age \leq 29.250 \;\&\; \nonumber\\
& is\_violent\_recid = 1)\nonumber\ { target} \Rightarrow {High}  \} \nonumber
\end{align}

\begin{center}
	\begin{tabular}{ cccc } 
		%	\toprule
		\multicolumn{4}{c}{	{\text{$\Re^+_{{c\alpha\_}{{\text{LoRMIkA}}}}$}}}\\
		\midrule
		%\hline
		Support = 0.226 & Coverage = 0.24 & Confidence = 0.93 & Lift= 1.22
	\end{tabular}
	
\end{center}

%\textit{Results:}{ 
The instance is being predicted as a prisoner with ``High'' likelihood to re-offend since the age is greater than or equal to 26 and less than or equal to 29.25, and this prisoner is a violent recidivist. %}
%The actual characteristics of the prisoner are an age of 27 and a classification as violent recidivist, which is consistent with the current supporting rule of LoRMIkA. 
%
Anchor and LORE produce the following rules to explain the same instance:

\begin{align}
\Re^+_{{c\alpha\_}{{\text{Anchor}}}} &=  \{( length\_of\_stay > 8.00 \;\& \nonumber\ two\_year\_recid = 1 \;\&\; 
race = African-American \;\&\; \nonumber\\
&is\_recid = 1 \;\&\;
c\_charge\_degree = F \;\&\;\nonumber\
sex = Male \;\&\; \nonumber\\
& days\_b\_screening\_arrest \leq 1.00 \; 
\&\; 25 < age \leq 31 \;\&\; 
 2 < priors\_count \leq 5 \;\&\; \\
 &is\_violent\_recid = 1 )\nonumber\ {target} \Rightarrow {High}\}   \nonumber
\end{align}

\begin{equation*}
\Re^+_{{c\alpha\_}{{\text{LORE}}}} = \nonumber\ \{(  age \leq 27  )\ \nonumber{target} \Rightarrow {High}\}\nonumber
\end{equation*}

LoRMIkA's current supporting rule shares some similarity with Anchor and LORE. We find that ($\Re^+_{{c\alpha\_}{{\text{LoRMIkA}}}} $) indicates that the prisoner has a high likelihood to re-offend if $26 \leq age \leq 29.250$. The supporting rules of LORE ($\Re^+_{{c\alpha\_}{{\text{LORE}}}} $) indicate $age \leq 27$, and Anchor ($\Re^+_{{c\alpha\_}{{\text{Anchor}}}} $) indicates $25 < age \leq 31$.

However, we argue that LoRMIkA's current supporting rule in this example is easier to understand and more informative than the rules provided by Anchor and LORE. Following \citet{Nauck2002-cl}, rules with a fewer number of conditions can be considered more interpretable than rules with a higher number of conditions. 
LORE ($\Re^+_{{c\alpha\_}{{\text{LORE}}}}$) produces a single condition with the ``age'' feature. Though we can consider this as highly interpretable, it seems not enough to decide whether the prisoner is a ``High'' level risk case to re-offend. The current supporting rule of LoRMIkA ($\Re^+_{{c\alpha\_}{{\text{LoRMIkA}}}} $) with two conditions is also brief and intuitive to understand, but it also can be considered highly related and relevant to the prediction in this question, as the other condition of $\Re^+_{{c\alpha\_}{{\text{LoRMIkA}}}}$ is $is\_violent\_recid = 1$.
Anchor ($\Re^+_{{c\alpha\_}{{\text{Anchor}}}} $) on the other hand provides a rather verbose rule with 11 conditions, which includes the conditions from the LoRMIkA rule, but points narrowly to the outcome of the particular decision, and seems harder to interpret due to its complexity.

\paragraph{What are the current conditions that contradict the prediction as having a High chance to re-offend?} As an answer to this question, we use the current contradicting rules ($\Re^-_{c\beta}$).  Below, we present the generated current contradicting rule of LoRMIkA. To the best of our knowledge, LoRMIkA is the only algorithm that provides current contradicting rules for the prediction. 

%\begin{comment}
%
%\begin{align}
%\Re^+_{{c\beta\_}{\textbf{\text{LoRMIkA}}}} &=  \{( priors\_count < 4.520 \;\&\; )\nonumber\  \textbf{target} \Rightarrow \textbf{Medium\_low}  \} \nonumber
%\end{align}
%\end{comment}

\begin{align}
\Re^-_{{c\beta\_}{{\text{LoRMIkA}}}} &=  \{( priors\_count < 5)\nonumber\ {target} \Rightarrow {Medium\_low}  \} \nonumber
\end{align}
\begin{center}
	\begin{tabular}{ cccc } 
		%	\toprule
		\multicolumn{4}{c}{	{\text{$\Re^-_{{c\beta\_}{{\text{LoRMIkA}}}}$}}}\\
		\midrule
		%\hline
		Support = 0.18 & Coverage = 0.331 & Confidence = 0.546 & Lift= 2.31 \\ \nonumber\\	
	\end{tabular}	
\end{center}

The rule can be used by decision-makers as suggestion which features should maintain their values to achieve a prediction of Medium\_Low likelihood to re-offend in the future.
%should adhere to the contradicting rule when making decisions to classify this prisoner as a Medium\_Low likelihood to re-offend (i.e., $priors\_count < 5$). 
In our example, the rule suggests that if this prisoner will re-offend at least one more time he will lose the only protective factor which potentially supports a prediction of low likelihood to re-offend.
The actual characteristic of this prisoner is \{$priors\_count = 4$)\}, so this can be interpreted in a way that this prisoner is at a tipping point of being a regular and frequent offender.

We note that the rule has a high lift (2.31), but fairly low confidence (0.546) which determines that this rule is not highly predictive, but can be seen as highly interesting.
Anchor and LORE do not consider $priors\_count$ of the prisoner for their current supporting rules, so that this information is also implicitly there in those algorithms, but our algorithm is able to make it explicit.%, by not only focusing on confidence.
 
%Proofs for the accuracy of LoRMIkA's current contradicting rule: Here, we validate the accuracy of LoRMIkA's current contradicting rule by examining the output of Anchor and LORE's current supporting rule. We assume that there is a high chance to be a current contradicting rule if the conditions in the antecedent of the current contradicting rule in LoRMIkA are not present in the antecedent of the current supporting rule of Anchor or LORE while the consequent remains same. In this example, Anchor and LORE do not consider priors\_count of the prisoner when providing the explanation for the current supporting rule, which is consistent with our argument. This is a sufficient evidence to prove that LoRMIkA's current contradicting rule (i.e., $priors\_count < 5$) is accurate. 

\paragraph{What are the hypothetical conditions that support the prediction as having a High chance to re-offend?} Below, we present the generated hypothetical supporting rule of LoRMIkA ($\Re^+_{h\alpha}$). To the best of our knowledge, LoRMIkA is the only algorithm that provides hypothetical supporting rules for a prediction.  
\begin{align}
\Re^+_{{h\alpha\_}{{\text{LoRMIkA}}}} &=  \{( priors\_count > 6 \;\&\; )\nonumber\  {target} \Rightarrow {High}  \} \nonumber
\end{align}

\begin{center}
	\begin{tabular}{ cccc } 
		%	\toprule
		\multicolumn{4}{c}{	{\text{$\Re^+_{{h\alpha\_}{{\text{LoRMIkA}}}}$}}}\\
		\midrule
		%\hline
		Support = 0.332 & Coverage = 0.33 & Confidence = 0.98 & Lift= 1.29 \\ \nonumber\\	
	\end{tabular}
\end{center}
 
 The rule is in line with the current contradicting rule, emphasizing that a further increase of $priors\_count$ would very negatively affect predicted risks for this prisoner in particular.
% change in the 
%	Possessing a priors count of more than 6 will increase the risk of a high chance of re-offending. The authorities should be aware of the chances of increasing the risk of re-offending this prisoner.
%
%Therefore, the relevant authorities should avoid releasing this prisoner when the priors count is more than 6. Further, authorities should impose rules and warn the prisoner that the offender will not be released to the society if the offender follows the same behaviour further since the offender  is already in his margin of the priors count. 
%
%The actual characteristic of this prisoner is \{$priors\_count = 4$)\}, meaning that the priors\_count of this prisoner is less than 6, which is not yet fulfilled by this prisoner. Therefore this generated rule of LoRMIkA can be classified as a hypothetical supporting rule for the predictive class.

\paragraph{What are the hypothetical conditions that could potentially invert the prediction as having a high chance to re-offend to medium\_low chance to re-offend?} As a solution for this question, we use the hypothetical contradicting rules ($\Re^-_{h\beta}$), which are often also called counterfactuals. Below, we present the generated hypothetically contradicting rules of LoRMIkA and LORE. 
\begin{align}
\Re^-_{{h\beta\_}{{\text{LoRMIkA}}}} &=  \{(age > 29 \;\&\; priors\_count < 4 )\nonumber\ {target} \Rightarrow {Medium\_Low}  \} \nonumber
\end{align}

\begin{center}
	\begin{tabular}{ cccc } 
		%	\toprule
		\multicolumn{4}{c}{	{\text{$\Re^-_{{h\beta\_}{{\text{LoRMIkA}}}}$}}}\\
		\midrule
		%\hline
		Support = 0.098 & Coverage = 0.11 & Confidence = 0.87 & Lift= 3.7 	
	\end{tabular}	
\end{center}
\begin{align}
\Re^-_{{h\beta\_}{{\text{LORE}}}} &=  \{( days\_b\_screening\_arrest > 0 \;\&\; age >28 )\nonumber\  {target} \Rightarrow {Medium\_Low}  \} \nonumber
\end{align}

	If this prisoner was older than 29 and $priors\_count$ was less than 4, this person would be identified as a prisoner with ``Medium\_Low'' likelihood to re-offend by the algorithm, according to LoRMIkA.
The actual characteristics of this prisoner are \{$age=27$, $priors\_count=4$\}.
Though $priors\_count$ in practice cannot be reduced and therewith the rule is not actionable, together with the other rules from above, the explanations suggest a prisoner at a tipping point, in terms of age and prior offences. The explanations suggest that if decision-makers can achieve that this prisoner will not re-offend in the next 2 years (approx.), the predicted risk will be lower afterwards.
%
%The hypothetical negative rule indicates that if \{$age > 29 \nonumber\ \& \nonumber\ priors\_count < 4$\} could reverse the prediction of the prisoner with ``Medium\_Low'' likelihood to re-offend, thus these rules show correlations that may not necessarily be causal. 
%
LoRMIkA's hypothetically contradicting rule shares some similarity with LORE, in terms of the $age$.

\subsection{Quantitative analysis}

In this section we present a quantitative analysis of the efficiency, stability, and interpretability of the rules produced by the different methods.

\subsubsection{Analysis of the efficiency of the rules produced by LoRMIkA compared with the rules of the other state-of-the-art algorithms}
%%%%%%%%%%%%%%%%%%%%%%%%%%%%%%%%
Following the key assumption of the local interpretability theory~\citep{ribeiro2016should, Ribeiro2018-mc, DBLP:journals/corr/abs-1805-10820}, we investigate whether the rule-based explanation produced for a particular instance is homogeneous across its neighbourhood using the association rule mining measures (i.e., coverage, confidence, and lift).
%\textbf{TODO: Not sure what this means}

The accuracy and the efficiency of the local rule-based model-agnostic explainers presented here is based on 1) the coverage that measures the fraction of instances in the neighbourhood which satisfy the LHS of the rule, 2) the confidence (i.e., precision, strength) which measures the percentage of the instances which satisfy the RHS of the rule, out of the instances selected for the coverage and 3) the lift which measures the interestingness of the rule.

Moreover, we compute the average time taken to generate a single explanation
for a prediction produced by each global machine learning model.  
%Fidelity compares the predictions of the explainer model and the blackbox model on the neighbourhood. 

\begin{table}[htbp]
	\caption{Coverage level of explainers} % title
	\label{tab:Coverage}
	% used for centering table
	\resizebox{\textwidth}{!}{%
		\begin{tabular}{|c|ccc|ccc|ccc|ccc|}
			\toprule
			{Dataset} &  \multicolumn{3}{|c|}{COMPAS} & \multicolumn{3}{|c|}{Adult} & \multicolumn{3}{|c|}{German}& \multicolumn{3}{|c|}{Covertype} \\
			\midrule
			Black-box & Anchor & LORE & LoRMIkA & Anchor & LORE & LoRMIkA & Anchor & LORE & LoRMIkA& Anchor & LORE & LoRMIkA\\ \hline
			SVM & 0.24 
			$\mypm$0.10 & 0.50
			$\mypm$0.12 & \textbf{0.65}  
			$\mypm$\textbf{0.13} & 0.09 
			$\mypm$0.09 & 0.50 
			$\mypm$0.10 & \textbf{0.73} 
			$\mypm$\textbf{0.12} & 0.27 
			$\mypm$0.13 & \textbf{0.55}
			$\mypm$\textbf{0.09} & 0.37
			$\mypm$0.03 & 	- 
			 & -
			 & -
			\\ 
			DT & 0.05 
			$\mypm$0.05 & 0.51 
			$\mypm$0.13& \textbf{0.79} 
			$\mypm$\textbf{0.12} & 0.07 
			$\mypm$0.09 & 0.49 
			$\mypm$0.11 & \textbf{0.83}
			$\mypm$\textbf{0.05} & 0.03 
			$\mypm$0.03 & 0.50 
			$\mypm$0.12 & \textbf{0.60} 
			$\mypm$\textbf{0.15} 	& 0.00 
			$\mypm$0.00 & 0.30
			$\mypm$0.18 & \textbf{0.92}
			$\mypm$\textbf{0.14} \\ 
			LR & 0.18 
			$\mypm$0.13 & 0.52 
			$\mypm$0.14 & \textbf{0.88}
			$\mypm$\textbf{0.05} & 0.10 
			$\mypm$0.10 & 0.49 
			$\mypm$0.11 & \textbf{0.88} 
			$\mypm$\textbf{0.03} & 0.06 
			$\mypm$0.07 & 0.51 
			$\mypm$0.10 & \textbf{0.77}
			$\mypm$\textbf{0.14} 	& 0.00
			$\mypm$0.00 & 0.41
			$\mypm$0.04 & \textbf{0.96}
			$\mypm$\textbf{0.09} \\ 
			NN & 0.09 
			$\mypm$0.08 & 0.50 
			$\mypm$0.13 & \textbf{0.84} 
			$\mypm$\textbf{0.08} & 0.05 
			$\mypm$0.06 & 0.49 
			$\mypm$0.10 & \textbf{0.86} 
			$\mypm$\textbf{0.03 }& 0.03 
			$\mypm$0.04 & 0.51 
			$\mypm$0.10 & \textbf{0.68}
			$\mypm$\textbf{0.17} 	& 0.00
			$\mypm$0.00 & 0.67
			$\mypm$0.41 & \textbf{0.96}
			$\mypm$\textbf{0.09} \\  
			RF & 0.08 
			$\mypm$0.07 & 0.49
			$\mypm$0.11 & \textbf{0.82} 
			$\mypm$\textbf{0.11} & 0.09 
			$\mypm$0.11 & 0.49 
			$\mypm$0.10 & \textbf{0.83} 
			$\mypm$\textbf{0.05} & 0.19 
			$\mypm$0.17 & 0.51 
			$\mypm$0.10 & \textbf{0.70} 
			$\mypm$\textbf{0.08} 	& 0.00
			$\mypm$0.00 & 0.36
			$\mypm$0.13 & \textbf{0.97}
			$\mypm$\textbf{0.09} \\ 
			\bottomrule
		\end{tabular}
	}
	% is used to refer this table in the text
\end{table}

\begin{table}[htbp]
	\caption{Confidence of explainers} % title of Table
	\label{tab:Precision} 
	% used for centering table
	\resizebox{\textwidth}{!}{%
		\begin{tabular}{|c|ccc|ccc|ccc|ccc|}
			\toprule
			{Dataset} &  \multicolumn{3}{|c|}{COMPAS} & \multicolumn{3}{|c|}{Adult} & \multicolumn{3}{|c|}{German} & \multicolumn{3}{|c|}{Covertype}\\
			\midrule
			Black-box & Anchor & LORE & LoRMIkA & Anchor & LORE & LoRMIkA & Anchor & LORE & LoRMIkA &Anchor & LORE & LoRMIkA \\ \hline
			SVM & 0.99
			$\mypm$0.01 & 0.98
			$\mypm$0.14 & \textbf{0.99}
			$\mypm$\textbf{0.01} & 0.98
			$\mypm$0.02 & 0.98
			$\mypm$0.05 & \textbf{1.00}
			$\mypm$\textbf{0.00} & 0.98
			$\mypm$0.01 & 0.99
			$\mypm$0.00 & \textbf{0.99}
				$\mypm$\textbf{0.00} 	& 
			-& -
				 & -
				\\ 
			DT & \textbf{0.98}
			$\mypm$\textbf{0.02} & 0.97
			$\mypm$0.08 & 0.91
			$\mypm$0.03 & 0.97
			$\mypm$0.02 & \textbf{0.98}
			$\mypm$\textbf{0.05} & 0.96
			$\mypm$0.01 & \textbf{0.98}
			$\mypm$\textbf{0.02} & 0.98
			$\mypm$0.06 & 0.84
			$\mypm$0.04 	& \textbf{0.88}
			$\mypm$\textbf{0.08} & 0.68
			$\mypm$0.41 & 0.80
			$\mypm$0.05 \\  
			LR & 0.97
			$\mypm$0.02 & 0.97
			$\mypm$0.09 & \textbf{1.00}
			$\mypm$\textbf{0.00} & 0.97
			$\mypm$0.01 & 0.98
			$\mypm$0.04 & \textbf{1.00}
			$\mypm$\textbf{0.00} & 0.98
			$\mypm$0.01 & 0.98
			$\mypm$0.06 & \textbf{0.99}
			$\mypm$\textbf{0.01}	& \textbf{0.97 }
			$\mypm$\textbf{0.01} & 0.93
			$\mypm$0.13 & \textbf{0.97}
			$\mypm$\textbf{0.01} \\ 
			NN & 0.98
			$\mypm$0.02 & 0.97
			$\mypm$0.08 & \textbf{0.99}
			$\mypm$\textbf{0.01} & 0.97
			$\mypm$0.01 & 0.98
			$\mypm$0.05 & \textbf{1.00}
			$\mypm$\textbf{0.00} & 0.98
			$\mypm$0.02& 0.98
			$\mypm$0.06 & \textbf{1.00}
			$\mypm$\textbf{0.00} 	& \textbf{0.96}
			$\mypm$\textbf{0.02} & 0.67
			$\mypm$0.40 & 0.88
			$\mypm$0.03 \\ 
			RF & 0.97
			$\mypm$0.02 & 0.98
			$\mypm$0.05 & \textbf{0.98}
			$\mypm$\textbf{0.01} & 0.97
			$\mypm$0.01 & 0.98
			$\mypm$0.05 & \textbf{0.99}
			$\mypm$\textbf{0.01} & 0.97
			$\mypm$0.01 & \textbf{0.98}
			$\mypm$\textbf{0.05} & 0.97
			$\mypm$0.01 	& 0.93
			$\mypm$0.06 & 0.80
			$\mypm$0.30 & \textbf{0.95}
			$\mypm$\textbf{0.02} \\  
			\bottomrule
		\end{tabular}
	}
	% is used to refer this table in the text
\end{table}

\begin{table}[htb]
	\caption{Rate of Interestingness}
	\label{tab:Lift_dif} 
	% used for centering table
	\begin{center}
		\resizebox{\textwidth}{!}{%
		\begin{tabular}{|c|ccc|ccc|ccc|ccc|}
	\toprule
	{Dataset} &  \multicolumn{3}{|c|}{COMPAS} & \multicolumn{3}{|c|}{Adult} & \multicolumn{3}{|c|}{German}& \multicolumn{3}{|c|}{Covertype} \\
	\midrule
	Black-box & Anchor & LORE & LoRMIkA & Anchor & LORE & LoRMIkA & Anchor & LORE & LoRMIkA& Anchor & LORE & LoRMIkA\\ \hline
				SVM & 0.01
				$\mypm$0.01	& 0.83
				$\mypm$0.32	& \textbf{3.11}
				$\mypm$\textbf{0.18}& 0.02
				$\mypm$0.02	& 0.84
				$\mypm$0.29	& \textbf{2.61}
				$\mypm$\textbf{0.94}	& 0.02
				$\mypm$0.01	& 0.90
				$\mypm$0.09	& 1.90
				$\mypm$0.17		& 
				- & -
				 & - \\  
				DT & 0.02 
				$\mypm$0.02 & 0.82 
				$\mypm$0.34 & \textbf{1.25}
				$\mypm$\textbf{0.17}& 0.03
				$\mypm$0.02 & 0.81
				$\mypm$0.34 & \textbf{1.85} 
				$\mypm$\textbf{0.61} & 0.02
				$\mypm$0.02 & 0.82
				$\mypm$0.35 & \textbf{0.99}
				$\mypm$\textbf{0.19}	& 1.00 
				$\mypm$0.00 & 0.50
				$\mypm$0.84 & \textbf{1.58}
				$\mypm$\textbf{0.20} \\ 
				LR & 0.03
				$\mypm$0.02&  0.81
				$\mypm$0.33& \textbf{3.33}
				$\mypm$\textbf{0.49} & 0.03
				$\mypm$0.01 & 0.82
				$\mypm$0.34 & \textbf{5.14}
				$\mypm$\textbf{0.98} & 0.02
				$\mypm$0.01 & 0.83
				$\mypm$0.30& \textbf{3.16}
				$\mypm$\textbf{0.58}	& 1.00
				$\mypm$0.00 & 0.98
				$\mypm$0.02 & \textbf{2.05}
				$\mypm$\textbf{0.12} \\ 
			NN & 0.02
		$\mypm$0.02 & 0.82
		$\mypm$0.33 & \textbf{2.87}
		$\mypm$\textbf{0.44} & 0.03
		$\mypm$0.01 & 0.83
		$\mypm$0.33 &\textbf{5.22}
		$\mypm$\textbf{1.24} & 0.02
		$\mypm$0.02& 0.88
		$\mypm$0.29 & \textbf{2.43}
		$\mypm$\textbf{0.29} 	& 1.00
		$\mypm$0.00 & 0.48
		$\mypm$0.84 & \textbf{1.90}
		$\mypm$\textbf{0.39} \\  
		RF & 0.03
		$\mypm$0.02& 0.81
		$\mypm$0.34 & \textbf{2.05}
		$\mypm$\textbf{0.31} & 0.03
		$\mypm$0.01 & 0.82
		$\mypm$0.35 & \textbf{3.19}
		$\mypm$\textbf{1.20} & 0.03
		$\mypm$0.01 & 0.83
		$\mypm$0.33 & \textbf{2.17}
		$\mypm$\textbf{0.66}	& 1.00 
		$\mypm$0.00 & 0.78
		$\mypm$0.59 & \textbf{2.02}
		$\mypm$\textbf{0.40} \\ 
		\bottomrule
			\end{tabular}
		}
	\end{center}
	% is used to refer this table in the text
\end{table}

\begin{table}[htbp]
		\caption{Run-time of explainers}
	\label{tab:Time} 
	\begin{center}
		\resizebox{0.45\textwidth}{!}{%
			\begin{tabular}{ |c|c|c|c|  }
			
			\hline
			\multicolumn{4}{|c|}{Covertype Dataset} \\
			\hline
			Black-box& Anchor &  LORE& LoRMIkA\\
			\hline
			DT   & 41.28s    &\textbf{22.60s}&   38.31s\\
			LR&   \textbf{2.19s } & 7.42s   &38.15s\\
			NN &20.48s & \textbf{20.13s}&  38.29s\\
			RF    &  \textbf{24.81s} & 289.41s&  38.27s\\
			
			\hline
		\end{tabular}
	}
	\end{center}

\end{table}

\paragraph{Coverage} Table~\ref{tab:Coverage} shows the mean values for the coverage of the rules generated as the explanations of LoRMIkA, Anchor, and LORE for the COMPAS, Adult, German and Covertype datasets. The best values across the local explainers are indicated with a boldface font.
%In LoRMIkA, 82\%, 83\%, 68\% of instances are supported by the rule-based explanations in COMPAS, Adult, and German datasets respectively, using the median from all the global models.  This suggests  that LoRMIkA outperforms LORE and Anchor local rule-based model-agnostic explainers.
%The median coverages over all the  global models  on LoRMIkA, Anchor, and LORE, respectively,
%are 82\%, 9\%, and 50\%
%for the COMPAS dataset, 83\%, 9\%, 49\% for the Adult dataset and 68\%, 6\%, 51\% for the German dataset.
When comparing the mean coverages of the rules over all the global models, LoRMIkA outperforms LORE and Anchor across all datasets and models, except one case, namely for the SVM model in the German dataset, where LORE performs better.
We argue that this is the case due to the flexibility of LoRMIkA to search and obtain k-optimal rules with high coverage.
As coverage measures how representative a rule is for a given dataset, our results show that LoRMIkA achieves the most representative rules.

%Table \ref{tab:Coverage},Table \ref{tab:Precision} and Table \ref{tab:Lift} illustrate that LoRMIkA maintains the highest coverage, precision and the lift in most of the instances.
\paragraph{Confidence} Table~\ref{tab:Precision} shows the mean values for the confidence of the rules generated as the explanations of LoRMIkA, Anchor and LORE for COMPAS, Adult and German datasets.
% In LoRMIkA, at the median of,  99\%, 1\%, 99\% of instances are supported by the antecedent and the consequent of the rule-based explanations in COMPAS, Adult, and German datasets respectively, considering all the global models (i.e., SVM, DT, LR, NN, RF) suggesting that LoRMIkA outperforms LORE and Anchor local rule-based model-agnostic explainers.
%  The median confidence over all the  global models is 99\%, 98\%, and 97\% for the COMPAS dataset, 100\%, 97\%, 98\% for the Adult dataset and 99\%, 98\%, 98\% for the German dataset, from LoRMIkA, Anchor, and LORE, respectively.
%When comparing the mean confidence of the rules over all the global models,
We see that LoRMIkA outperforms LORE and Anchor in most instances. In particular, it outperforms the comparison methods in all instances but the DT instances across all datasets, the NN in the Covertype dataset and the RF instance in the German dataset.
Our results show that LORE and Anchor provide rule-based explanations with competitive mean confidence values since their main focus is to find the rules with the highest confidence. However, LoRMIkA is able to achieve the highest mean confidence in most cases. 

\paragraph{Lift} %Table~\ref{tab:Lift} shows the mean values for the lift of the rules generated as the explanations of LoRMIkA, LORE, and Anchor for the COMPAS, Adult, and German datasets. 
Table~\ref{tab:Lift_dif} shows the rate of interestingness which is calculated from the lift as an absolute difference from $1$ (as a lift of $1$ shows there is no association between antecedent and consequent).

%was calculated using Table~\ref{tab:Lift} based on the fact that how far the lift value deviated from 1 .
 
We note that the lift for Anchor is identical to its confidence, as it defines the neighbourhood in a way that it only contains positive instances. This way, the global model predictions of the instances in the neighbourhood are equal to the global model prediction of the instance to be explained. Per the definition of lift in Definition~\ref{Def:Lift} the $Support(q)$ is equal to $1$.  Therewith, in Anchor the lift is equal to the confidence according to Definition~\ref{Def:Strength}. Thus, lift for Anchor is by definition between $0$ and $1$, as well as the rate of interestingness.

We can see from the table that LoRMIkA outperforms LORE and Anchor in all instances but one, which is the SVM in the German dataset for LORE.
This again shows the flexibility of our algorithm LoRMIkA that makes it possible to obtain interesting rules with high lift and does not limit us to rules with the highest predictive power.

\paragraph{Computational Time}
The experiments are run on an Intel(R) i7 processor (3.2 GHz), with a single thread per core, and 64GB of main memory.
Table~\ref{tab:Time} shows the average running time over 50 instances measured in seconds, to produce explanations of the predictions generated by each global model for the Covertype dataset. 
From the table, we can see that Anchor runs fastest as an explainer for the LR and RF global models, whereas LORE runs the fastest for DT and NN. LoRMIkA has a very consistent running time around 38s across all the models.

\bigskip
Overall, our analysis shows that in most cases the efficiency of the rule-based explanations of LoRMIkA is superior to the LORE and Anchor baselines with respect to coverage, confidence, and rate of interestingness of the rules when explaining the predictions of the global models of COMPAS, Adult, German and Covertype datasets. With regards to the computational time, Anchor and LORE can be both considerably faster and considerably slower than LoRMIkA, while the latter one has a quite constant computation time.

\subsubsection{Analysis of the stability of the rules produced by LoRMIkA compared with the rules of the other state-of-the-art algorithms}

\begin{table}[htbp]
	\caption{Jaccard measure of stability in explainers} % title of Table
	\label{tab:Jaccard} 
	% used for centering table
	\resizebox{\linewidth}{!}{%
				\begin{tabular}{|c|ccc|ccc|ccc|ccc|}
			\toprule
			{Dataset} &  \multicolumn{3}{|c|}{COMPAS} & \multicolumn{3}{|c|}{Adult} & \multicolumn{3}{|c|}{German}& \multicolumn{3}{|c|}{Covertype} \\
			\midrule
			Black-box & Anchor & LORE & LoRMIkA & Anchor & LORE & LoRMIkA & Anchor & LORE & LoRMIkA& Anchor & LORE & LoRMIkA\\ \hline
			SVM & 0.85
			$\mypm$0.18 & 0.50
			$\mypm$0.12 & \textbf{1.00}
			$\mypm$\textbf{0.00}& 0.73
			$\mypm$0.25 & 0.91
			$\mypm$0.11 &  \textbf{1.00}
			$\mypm$\textbf{0.00} & 0.74
			$\mypm$0.26 & 0.93
			$\mypm$0.08 &  \textbf{1.00}
			$\mypm$\textbf{0.00} 	& 0.27 
			$\mypm$0.13 & \textbf{0.55}
			$\mypm$\textbf{0.09} & 1.00
			$\mypm$0.00 \\ 
			DT & 0.78
			$\mypm$0.19 & 0.74
			$\mypm$0.26 & \textbf{1.00}
			$\mypm$\textbf{0.00}& 0.75
			$\mypm$0.20 & 0.79
			$\mypm$0.24 &  \textbf{1.00}
			$\mypm$\textbf{0.00} & 0.74
			$\mypm$0.69 & 0.79
			$\mypm$0.24 &  \textbf{1.00}
			$\mypm$\textbf{0.00} 	& 0.63 
			$\mypm$0.22 & \textbf{0.74}
			$\mypm$\textbf{0.21} & 1.00
			$\mypm$0.00 \\  
			LR & 0.80
			$\mypm$0.18 & 0.52
			$\mypm$0.14 &  \textbf{1.00}
			$\mypm$\textbf{0.00} & 0.73
			$\mypm$0.26 & 0.82
			$\mypm$0.20 & \textbf{1.00}
			$\mypm$\textbf{0.00}& 0.74
			$\mypm$0.26 & 0.82
			$\mypm$0.21 & \textbf{1.00}
			$\mypm$\textbf{0.00} 	& 0.84 
			$\mypm$0.16 & \textbf{0.76}
			$\mypm$\textbf{0.21} &  1.00
			$\mypm$0.00 \\ 
			NN & 0.74
			$\mypm$0.24 & 0.83
			$\mypm$0.19 &  \textbf{1.00}
			$\mypm$\textbf{0.00} & 0.74
			$\mypm$0.22 & 0.78
			$\mypm$0.25 & \textbf{1.00}
			$\mypm$\textbf{0.00}& 0.68
			$\mypm$0.26 & \textbf{1.00}
			$\mypm$\textbf{0.00} & \textbf{1.00}
			$\mypm$\textbf{0.00} 		& 0.66
			$\mypm$0.23 & \textbf{0.76}
			$\mypm$\textbf{0.19} &  1.00
			$\mypm$0.00 \\ 
			RF & 0.79
			$\mypm$0.23 & 0.49
			$\mypm$0.11 &   \textbf{1.00}
			$\mypm$\textbf{0.00} & 0.74
			$\mypm$0.23 & 0.82
			$\mypm$0.20 &  \textbf{1.00}
			$\mypm$\textbf{0.00} & 0.75
			$\mypm$0.22 & 0.75
			$\mypm$0.20 &  \textbf{1.00}
			$\mypm$\textbf{0.00} 		& 0.65 
			$\mypm$0.21 & \textbf{0.67}
			$\mypm$\textbf{0.26} & 1.00
			$\mypm$0.00 \\ 
			\bottomrule
		\end{tabular}
	}
	% is used to refer this table in the text
\end{table}

%\begin{comment}
%Both Anchor and LORE algorithms generate rules with low coverage and high
%precision. LoRMIkA in contrast has an adaptable optimisation criterion that can be changed according to the task at hand. For instance, when evaluating coverage we can obtain rules with the highest coverage. This shows the flexibility of our algorithm that makes it possible to obtain interesting and predictive rules and doesn't limit us to rules with the highest predictive power as the state-of-the-art algorithms we compare against. Moreover, in contrast to LORE, LoRMIkA provides interesting rules with high lift. 
%\end{comment}
%In Table \ref{tab:Fidelity} we do not consider Anchor, as it changes the neighbourhood of the rules using a bandit algorithm until the neighbourhood satisfies most of the criteria in the LHS of the rule. 

We evaluate the stability of the local rule-based model-agnostic explainers by calculating the similarity of the resulting rules over independent runs, using the Jaccard coefficient as defined in Equation~\ref{Eq:Jaccard}.

\begin{equation}\label{Eq:Jaccard}
J(X,Y)=|X \cap Y|/|X\cup Y|
\end{equation}

Here, $X$ and $Y$ are two given sets of features that are included in the rules from two runs. The Jaccard coefficient calculates the similarity~\citep{Kuznetsov2018-kc} by comparing the common and distinct features in the two sets. The output ranges from 0 to 1, and the higher the coefficient the higher the similarity of rules over the two runs.
In particular, we randomly select 50 instances and evaluate the stability of the generated explanations over 10 independent runs.

Table~\ref{tab:Jaccard} indicates the mean and the standard deviation of the Jaccard coefficient. When measuring the Jaccard coefficient, we consider rules generated for highest confidence and rules which support the current prediction (i.e., $\Re^+_{c\alpha}$). 
LoRMIkA produces the rules with the highest stability compared with LORE and Anchor with a Jaccard coefficient of 1.00 in all occasions, indicating the high stability of the method.
We argue that this high stability is due to the robust nature of our approach when selecting the neighbourhood as explained in Section~\ref{Sec:Approach}. 

%The median Jaccard coefficient  over all the defined global models (i.e., SVM, DT, LR, NN, RF) is 100\%, 80\% and 50\% for the COMPAS dataset, 100\%, 73\% and 82\%  for the Adult dataset and 100\%, 74\% and 82\%  for the German dataset, from LoRMIkA and LORE, respectively. 

% \begin{equation}\label{Eq:Jaccard}
% \end{equation}

%\textbf{TODO: I suppose you searched for coverage for the coverage table, for strength for the strength table, etc.?}
%\textbf{TODO: What is the Jaccard coefficient really telling us?}

\subsubsection{Quantitative analysis of interpretability}

%\subsubsection{Analyze the interpretability of the rules produced by LoRMIkA compared with the rules of the other state-of-the-art algorithms}

As the interpretability of a given method is inherently difficult to measure, as a proxy we use the simplicity of the generated explanations \citep{Bechlivanidis2017-xe}. We measure the total amount of features used in the antecedent of the given rule of the local rule-based explainers. 

\begin{table}[htbp]
	\caption{Number of features in the explanation} % title of Table
	\label{tab:Num_fea} 
	% used for centering table
	\resizebox{\linewidth}{!}{%
				\begin{tabular}{|c|ccc|ccc|ccc|ccc|}
			\toprule
			{Dataset} &  \multicolumn{3}{|c|}{COMPAS} & \multicolumn{3}{|c|}{Adult} & \multicolumn{3}{|c|}{German}& \multicolumn{3}{|c|}{Covertype} \\
			\midrule
			Black-box & Anchor & LORE & LoRMIkA & Anchor & LORE & LoRMIkA & Anchor & LORE & LoRMIkA& Anchor & LORE & LoRMIkA\\ \hline
			SVM & 1.71 
			$\mypm$0.25 & 1.36 
			$\mypm$0.29& \textbf{1.03} 
			$\mypm$\textbf{0.19}
			& 2.61 
			$\mypm$0.31 & 1.24 
			$\mypm$0.25& \textbf{1.00} 
			$\mypm$\textbf{0.00}& 2.87 
			$\mypm$0.37& 1.10 
			$\mypm$0.10& \textbf{1.00} 
			$\mypm$\textbf{0.00}	& -
			 & -
			  & - \\  
			DT & 4.77 
			$\mypm$0.47& 1.84 
			$\mypm$0.59& \textbf{1.25} 
			$\mypm$\textbf{0.44}& 4.23 
			$\mypm$0.58& 2.30 
			$\mypm$0.61& \textbf{1.26} 
			$\mypm$\textbf{0.45}& 4.27 
			$\mypm$1.05& 2.44 
			$\mypm$0.90& \textbf{1.21} 
			$\mypm$\textbf{0.42}		& 28.50 
			$\mypm$8.28 & 4.77
			$\mypm$0.77 & \textbf{2.25}
			$\mypm$\textbf{1.29} \\  
			LR & 2.08 
			$\mypm$0.31 & \textbf{1.28} 
			$\mypm$\textbf{0.30}& 1.40 
			$\mypm$0.50& 2.73 
			$\mypm$0.52& 1.90
			$\mypm$0.47& \textbf{1.41} 
			$\mypm$\textbf{0.50}& 2.80 
			$\mypm$0.45& 1.85 
			$\mypm$0.46& \textbf{1.20}
			$\mypm$\textbf{0.42} 		& \textbf{3.21} 
			$\mypm$\textbf{0.64} & 4.11
			$\mypm$0.74 & 6.46
			$\mypm$3.59 \\  
			NN & 3.99 
			$\mypm$0.47& \textbf{1.32} 
			$\mypm$\textbf{0.29}& 1.38 
			$\mypm$0.53& 3.70 
			$\mypm$0.39& 1.98 
			$\mypm$0.49& \textbf{1.46} 
			$\mypm$\textbf{0.57}& 4.03 
			$\mypm$0.67 & \textbf{1.00} 
			$\mypm$\textbf{0.00}& 1.27 
			$\mypm$0.45			& 10.00 
			$\mypm$3.68 & 6.13
			$\mypm$0.96 & \textbf{5.48}
			$\mypm$\textbf{3.78} \\ 
			RF & 4.20 
			$\mypm$0.34& 1.61 
			$\mypm$0.43& \textbf{1.37} 
			$\mypm$\textbf{0.50}& 3.46 
			$\mypm$0.45& 2.04 
			$\mypm$0.47& \textbf{1.28} 
			$\mypm$\textbf{0.44}& 3.10 
			$\mypm$0.60& 2.50 
			$\mypm$0.91 & \textbf{1.07} 
			$\mypm$\textbf{0.28}		& 15.39 
			$\mypm$4.36 & \textbf{3.60}
			$\mypm$\textbf{0.80} & 8.77
			$\mypm$7.46 \\  
			\bottomrule
		\end{tabular}
	}
	% is used to refer this table in the text
\end{table}

Table~\ref{tab:Num_fea} indicates the mean values of the number of features used in a single rule to explain each instance using Anchor, LORE, and LoRMIkA. 
LoRMIkA outperforms LORE and Anchor across all datasets and models, except five cases, namely for the LR and NN models in the COMPAS dataset and the NN model in the German dataset, and LR and RF in the Covertype dataset where LORE performs better except LR in the Covertype dataset. 
This indicates that LoRMIkA is able to obtain more concise rules than Anchor and rules of comparable complexity to LORE. %Furthermore, in LoRMIkA, the maximum number of features can be set according to user preference. 

%\section{Future Works}
%\label{Sec:futureworks}
%This work is purely based on providing local rule-based explanations for the predictions produced by the machine learning models for the tabular classification datasets with defined attributes. Recently,~\citet{Zou2019-px} proposed a new approach to explain the predictions produced by the deep learning models in the recommendation.
%In their approach, they combine both latent features and the attributes to discover interpretable preferences.
%In the future, we hope to extend our algorithm to provide explanations for the predictions produced from datasets with latent variables.

\section{Conclusions}
\label{Sec:conc}

We have presented Local Rule-based Model Interpretability with k-optimal Associations (LoRMIkA), a model-agnostic framework for local explainability of classification algorithms.
It employs association rule mining techniques in a local neighbourhood to explain a particular instance and is therewith able to extract models that are local both with respect to features and instances. It uses the efficient OPUS search algorithm to extract top k-optimal rules with respect to measures such as support, coverage, confidence, lift, and leverage. This makes the framework flexible and allows us to extract simple rules that are not only predictive but also interesting and better suited to provide explanations.
The experiments performed have shown that LoRMIkA is competitive and able to outperform state-of-the-art local explainers both in quantitative and qualitative aspects. Moreover, in contrast to state-of-the-art local explainers, it is able to provide multiple rules which explain the prediction in various aspects. 

\section*{Acknowledgements}
\label{sec:ach}

%\noindent 
This research was supported by the Australian Research Council under grant DE190100045.

\section*{References}

\bibliographystyle{plainnat}
\bibliography{dilini2018lormika_insnew}

\end{document}